%% file: main.tex
\documentclass[11pt]{article}
\usepackage[letterpaper,margin=1in]{geometry}
\usepackage{amsmath,amssymb,amsfonts}
\usepackage{algorithmic}
\usepackage{algorithm}
\usepackage{graphicx}
\usepackage{textcomp}
\usepackage{xcolor}
\usepackage{booktabs}
\usepackage{colortbl}
\usepackage{authblk}
\usepackage{tikz}
\usetikzlibrary{arrows.meta,positioning,fit,backgrounds,calc,shapes.geometric}
\input{figures/figstyle}
\usepackage[hidelinks]{hyperref}

\newcommand{\code}[1]{\texttt{\expandafter\codebreak#1\relax}}
\makeatletter
\def\codebreak#1{%
  \ifx#1\relax\else
    #1%
    \ifx#1.\allowbreak\fi
    \ifx#1_\allowbreak\fi
    \expandafter\codebreak\fi}
\makeatother

\begin{document}

\title{\textbf{VST: Verifiable Structured Transport\\ for Auditable Agent-to-Agent Alpha Discovery}}

\author{\textbf{Yuqi Li} \quad \textbf{Siyuan Liu} \quad \textbf{Bingjun Liu}}
\affil{
  \textbf{Panda AI} \\
  \texttt{\{libubai, liusiyuan, liubingjun\}@pandaai.online}
}
\date{}

\maketitle

\begin{abstract}
Agent-to-agent (A2A) alpha discovery is slowed by repeated feedback cycles between mining and evaluation agents, whose hand-offs, in contemporary LLM multi-agent systems, are free-form natural-language messages that carry no stable contract and cannot be replayed. We first restructure this communication as a structured agent-to-agent protocol of \emph{typed, causally addressable, unicast records}, so that the committed stream forms a causal trajectory. On that trajectory a single predictor with four typed heads forecasts the accumulated guidance the two miners would receive several cycles ahead; a transactional verify--leap controller then commits a multi-cycle speculative outcome only when it passes a four-level gate, and otherwise rolls back to the exact prior state. Structure is the enabling contribution, and its value is not accuracy. A controlled ablation shows an equal-information free-text channel reaches the same predictor hit rate. What typing provides is a state that can be schema-checked, replayed deterministically, and prevented by construction from leaking a forecast to an evaluator: auditability by construction, not an empirically stress-tested guarantee. On a CSI~1000 out-of-sample holdout, our single run is the only one among eight methods (seven baselines and ours) to hold a positive median annualized return and Sharpe at the factor level, though the median return \emph{in excess} of the benchmark stays negative for every method including ours; its development-selected top-20 portfolios reach a $0.71$ median holdout Sharpe, selected on a split inside the optimization horizon. We report these single-run results descriptively, gross of costs, and are explicit about their limits throughout; in particular we do not isolate the effect of the leap machinery from the inherited search substrate, which we leave to future work.
\end{abstract}

\noindent\textbf{Keywords:} agent-to-agent communication, LLM multi-agent systems, speculative execution, alpha factor mining

\section{Introduction}
\input{sections/01_introduction}

\section{Related Work}
\input{sections/02_related_work}

\section{The VST Method}
\label{sec:method}

\subsection{System Setting and Problem Formulation}
\input{sections/03_formulation}

\subsection{Structured Inter-Agent Communication Protocol}
\input{sections/04_c1_structured_transport}

\subsection{Multi-Horizon Structured Trajectory Prediction}
\input{sections/05_c2_prediction}

\subsection{Transactional Verification and Leap Control}
\input{sections/06_c3_verify_leap}

\subsection{Separated Co-Evolution of Predictor and Evaluator Skills}
\input{sections/07_c4_coevolution}

\section{Experiments}
\input{sections/08_experiments}

\section{Discussion}
\input{sections/09_discussion}

\section{Limitations}
\label{sec:limitations}
\input{sections/10_limitations}

\section{Conclusion}
\input{sections/11_conclusion}







\bibliographystyle{IEEEtran}
\bibliography{references}

\appendix
\section{Aggregation Algorithm}
\label{app:agg}
\input{sections/appendix_aggregation}

\section{Operational Runtime Side Effects}
\label{app:runtime}
\input{sections/appendix_runtime}

\end{document}

%% file: figures/figstyle.tex
\definecolor{vspBlue}{HTML}{2C5AA0}
\definecolor{vspBlueBg}{HTML}{E8EEF7}
\definecolor{vspGreen}{HTML}{2E7D5B}
\definecolor{vspGreenBg}{HTML}{E4F0EA}
\definecolor{vspOrange}{HTML}{C56A1A}
\definecolor{vspOrangeBg}{HTML}{FBEBDA}
\definecolor{vspViolet}{HTML}{6A4C93}
\definecolor{vspVioletBg}{HTML}{EDE7F3}
\definecolor{vspRedBg}{HTML}{F8E3E3}
\definecolor{vspGray}{HTML}{555555}
\definecolor{vspGrayBg}{HTML}{EFEFEF}

\tikzset{
  vspfont/.style={font=\footnotesize},
  vspbox/.style={draw=vspGray, line width=0.5pt, rounded corners=2pt, align=center, inner sep=4pt},
  vsparrow/.style={-{Latex[length=2mm]}, line width=0.6pt, draw=vspGray},
  vspdash/.style={-{Latex[length=2mm]}, line width=0.6pt, draw=vspGray, dashed},
  vspL1/.style={vspbox, fill=vspBlueBg, draw=vspBlue},
  vspL2/.style={vspbox, fill=vspGreenBg, draw=vspGreen},
  vspL3/.style={vspbox, fill=vspOrangeBg, draw=vspOrange},
  vspL4/.style={vspbox, fill=vspVioletBg, draw=vspViolet},
  vspNeutral/.style={vspbox, fill=vspGrayBg},
  vspReject/.style={vspbox, fill=vspRedBg, draw=vspOrange},
}

\newcommand{\oursrow}{\rowcolor{vspBlueBg}}

%% file: sections/01_introduction.tex
Automated discovery of predictive signals, or \emph{alphas}, has moved from single-model symbolic search toward multi-agent pipelines in which several specialized components propose, evaluate, and refine candidates in concert. A representative design is a dual-loop evolutionary search rather than a linear chain, in which one report agent advises two miners: an alpha-mining loop proposes candidate factors and analyzes them with independent evaluator instances, while in parallel an evaluation-optimization loop proposes changes to the metric set and portfolio construction and analyzes the resulting configuration \cite{alphasingularity}. The two loops are coupled, because the evaluation side scores alpha candidates while accepted factors become evaluation-side samples. Only after the coupled state converges does the system commit the factor pool, the evaluation configuration, the report memory, and the approved skills. This dual-loop architecture is the substrate we build on; it is not a contribution of this paper. What we take from it is a concrete, demanding instance of a recurring problem: repeating these miner--evaluator--report cycles is the serial cost we seek to reduce.

The bottleneck lies in the number of round-trips between agents rather than in the computation inside any single agent. Each cycle waits for a miner to produce candidates, for several evaluators to return analyses, and for a report agent to integrate those analyses into both loops. When the search runs for hundreds of cycles, the wall-clock cost is dominated by this sequential feedback rather than by any one model call.

Contemporary LLM multi-agent frameworks show that role specialization and agent-to-agent (A2A) messaging can solve complex tasks \cite{hong2023metagpt,qian2024chatdev,wu2023autogen,li2023camel}. A recurring property of these systems, surveyed in \cite{guo2024survey}, is that agents coordinate primarily through \emph{free-form or loosely structured natural language}. Even where individual hand-offs are typed by structured-output schemas, the interaction as a whole is rarely exposed as a stable, replayable record contract, so the message stream cannot be aligned into a causal trajectory. These properties are acceptable when the only goal is to reach a task answer, but they foreclose two capabilities that matter for a long-running evolutionary search: predicting where the interaction is heading, and verifiably skipping ahead to that state.

Our central idea is that structuring the communication is a prerequisite, not an optimization. We reconstruct every inter-agent transmission as a typed three-layer record, an envelope that identifies and routes the message, a role-specific body that retains the original payload and receiver argument, and a transactional provenance block that records whether the message belongs to a committed or a speculative state (detailed in Section~\ref{sec:c1}). The wrapper gives each agent a stable input contract, removes ambiguous prose hand-offs, enforces visibility boundaries, and makes multi-cycle state transitions replayable.

Only once actual communication has this structure do its committed records form a trajectory that can be replayed and aligned into supervised targets. We then introduce one predictor that forecasts the accumulated \emph{miner-directed guidance} the two miners would receive over the next $k$ cycles, and a controller that runs a short speculative execution and commits the result only if it passes a layered validation. A prediction is never acted on directly: it proposes a candidate future state, and the real miners and evaluators then decide whether that state is valid. Structuring the channel is what makes this loop safe and auditable, because the forecast is schema-checked, its routing cannot leak to an evaluator, and a rejected leap rolls back to an exact prior state.

\paragraph{Contributions.}
Taking the dual-loop agentic alpha-discovery system of \cite{alphasingularity} as a fixed substrate, this paper turns its communication layer into a structured agent-to-agent protocol: a typed transport on which multi-cycle guidance can be forecast and a speculative leap can be verifiably committed. We call this paradigm VST (Verifiable Structured Transport), where the forecast is a prediction over the structured communication trajectory that is never trusted directly; it proposes a candidate future state that the real agents verify, and we make no claim that structuring improves forecast accuracy. The four contributions below follow from one idea: the inter-agent channel should be a typed, replayable record (C1). Once it is, three capabilities follow that a prose channel cannot support: the trajectory can be forecast (C2), a forecast can be verifiably committed or rolled back (C3), and the predictor can co-evolve without corrupting the evaluators that judge it (C4).
\begin{itemize}
  \item \textbf{C1: Structured inter-agent communication protocol.} A three-layer, A2A-compatible packet format with deterministic unicast routing and visibility typing that preserves the existing payload field paths and receiver arguments while making every transmission typed, causally addressable, and replayable (Section~\ref{sec:c1}). This is the enabling contribution. Its value is not accuracy: a controlled ablation (Section~\ref{sec:exp-ablation}) shows an equal-information free-text channel reaches the same predictor hit rate. What typing provides is a leak-proof, replayable trajectory at no cost to downstream effectiveness, an architectural correctness property the free-text channel cannot provide and on which the verified leap depends.
  \item \textbf{C2: Multi-horizon structured trajectory prediction.} A single agentic-RL predictor with four typed heads that jointly forecasts, for a chosen horizon $k \ge 2$, the accumulated guidance delivered to the two miners, with a calibrated success confidence and epistemic uncertainty (Section~\ref{sec:c2}). Two design choices are specific to our setting rather than inherited from standard LLM-agent RL: a single shared-encoder policy with four typed heads (not four independent agents), which keeps the coupled channels cross-consistent; and a transaction-level, verification-grounded reward, in which the four-level gate's commit/rollback outcome is the learning signal and rolled-back trajectories are retained as negatives (Section~\ref{sec:c4}).
  \item \textbf{C3: Transactional verification and leap control.} A four-level gate that validates the speculative \emph{outcome} rather than the forecast field-by-field, committing the complete transaction atomically or rolling it back to the last committed state (Section~\ref{sec:c3}).
  \item \textbf{C4: Separated skill co-evolution.} Distinct, transactionally promoted skill libraries for the predictor and the evaluators, which prevent a successful predictor from teaching the evaluators to agree with its own forecast (Section~\ref{sec:c4}).
\end{itemize}
The design borrows the verify-then-commit discipline of speculative decoding \cite{leviathan2023speculative} but applies it at the level of a multi-agent workflow: the unit of speculation is a multi-cycle transition of two coupled loops, not a token.

%% file: sections/02_related_work.tex
\subsection{LLM multi-agent systems and their communication layer}
We position our work against several literatures of unequal weight. How LLM agents communicate (this subsection) is where our load-bearing contribution lives; prediction and learning over agent trajectories, automated alpha discovery and financial LLM agents, and speculative execution are the traditions from which our downstream consequences borrow.
Role-specialized LLM agents now collaborate on multi-step tasks through a variety of coordination structures: fixed assembly-line workflows with standardized roles \cite{hong2023metagpt,qian2024chatdev}, general frameworks for composing conversational agents \cite{wu2023autogen,li2023camel}, and looser regimes in which agents debate, recompose their team, or simulate social behavior \cite{du2023debate,chen2023agentverse,park2023generative}; recent work scales this to networked topologies with a reported collaborative scaling law \cite{qian2025scaling}. What these designs share, and what several surveys now foreground as a first-class design axis \cite{guo2024survey,wang2024autonomous,xi2023rise,tran2025collaboration}, is that the coordination medium itself is natural language.

Across these designs the primary coordination medium is natural language. Modern
stacks do impose typed contracts on individual hand-offs (structured output,
JSON-mode, function/tool-calling), and systems such as MetaGPT exchange structured
intermediate artifacts rather than pure prose; what remains absent is not per-message
typing but a \emph{transactional, replayable record contract} over the whole
interaction. Typed performative envelopes that separate routing metadata from content
are themselves long-standing: agent communication languages such as KQML
\cite{finin1994kqml} and the FIPA-ACL tradition established them decades ago, and the
Model Context Protocol and Agent2Agent protocol revive them today. Our envelope sits
in this lineage; what we add to it is the \emph{transactional provenance}
layer (observed vs.\ speculative origin, commit status) and the visibility typing that
together make a speculative multi-cycle transaction replayable, leak-proof, and
rollback-safe, properties an ACL envelope alone does not provide. This also departs
from multi-agent reinforcement learning, where agents \emph{learn} a differentiable
communication protocol end-to-end \cite{foerster2016learning}, whereas contemporary
LLM agents exchange free-form text. Our work is orthogonal to the coordination
pattern: rather than propose a new role structure or prompting scheme, we replace the
untyped prose channel with a typed, replayable transport on which the same information
becomes a stable trajectory coordinate. Closest to our setting, a dual-loop
agent-to-agent architecture for self-evolving alpha discovery has been proposed
\cite{alphasingularity}, in which coupled mining and evaluation loops co-evolve through
repeated feedback. That work establishes the multi-agent search substrate; it
communicates through conventional hand-offs and does not treat the communication layer
itself as an object to be typed, forecast, and verifiably advanced, which is the gap
this paper addresses.

\subsection{Prediction and learning over agent trajectories}
Our predictor is an agentic reinforcement-learning component, and it forecasts a future state before acting on it, so it draws on two adjacent lines. The first turns an LLM into a policy: from prompting techniques that elicit or search over reasoning \cite{wei2022cot,yao2023tot}, through agents that interleave reasoning with tool use and self-critique \cite{yao2023react,schick2023toolformer,shinn2023reflexion,madaan2023selfrefine}, to policies trained with PPO-style RLHF and, more recently, group-relative objectives that incentivize reasoning directly \cite{schulman2017ppo,ouyang2022instructgpt,shao2024deepseekmath,deepseek2025r1}. Voyager's growing library of executable skills \cite{wang2023voyager} directly motivates our predictor skill library (Section~\ref{sec:c4}). These settings are single-agent and optimize toward an answer or action sequence; our reward is instead a \emph{transaction-level} signal from a coupled dual loop, and the predicted object is a multi-cycle guidance state (Section~\ref{sec:c4}). The second line is model-based control: latent world models that plan or imagine over a learned dynamics \cite{hafner2019planet,hafner2020dreamer}, and receding-horizon model predictive control that plans, executes, and re-anchors at the true state \cite{rawlings2017mpc}. VST plays the analogous role at the multi-agent level, with the committed communication trajectory as the observable dynamics; but where Dreamer trusts its imagined rollout for policy learning and MPC re-plans continuously over a numerical model, our forecast is never trusted directly, and verification is a discrete transactional gate with exact rollback.

\subsection{Automated alpha mining and financial LLM agents}
Formulaic alpha mining searches a combinatorial space of expression trees for return-predictive signals, using reinforcement learning to assemble complementary alpha collections \cite{yu2023alphagen}, symbolic regression with risk-seeking gradients \cite{petersen2021dsr}, LLM-assisted discovery \cite{yu2026hybrid}, agentic frameworks that regularize for decay-resistance and originality \cite{tang2025alphaagent}, and relational deep models for stock ranking \cite{feng2019temporal}. These target the quality of the discovered factors and treat the search as a monolithic optimization; the closest agentic baseline, AlphaAgent \cite{tang2025alphaagent}, couples an LLM agent with AST-level constraints but contributes a regularized generator. Ours is a communication substrate rather than a generator: we restructure the inter-agent channel and add a verified multi-cycle leap on top of an existing miner, treating the search as a multi-agent process whose inter-agent feedback is the dominant serial cost. The two directions are complementary, since our transport is agnostic to the underlying miner and could wrap an RL, symbolic, or LLM-based generator; our experiments use the Qlib platform \cite{yang2020qlib}. Beyond factor mining, LLMs have been applied broadly across finance \cite{li2023finllmsurvey}, from domain-specialized language models \cite{wu2023bloomberggpt,yang2023fingpt} and time-series forecasting \cite{yu2023temporal} to end-to-end trading via deep RL and multi-agent or multimodal decision systems \cite{liu2020finrl,zhang2024finagent,xiao2024tradingagents}. These operate at the trading-policy layer; our contribution sits below it, in how the research agents that produce signals communicate and evolve.

\subsection{Speculative execution}
Speculative decoding accelerates autoregressive generation by drafting several tokens with a cheaper model and verifying them against the target model, preserving the exact output distribution \cite{leviathan2023speculative}. VST adopts the same verify-then-commit principle but at a different granularity: the ``draft'' is a predicted multi-cycle guidance state, and ``verification'' is the actual execution of both miners and their blinded evaluators followed by a layered quality gate. The broader ``speculate optimistically, then validate and commit or abort'' pattern is not itself new: it is the defining discipline of optimistic concurrency control in databases \cite{kung1981optimistic} and of transactional memory in architecture \cite{herlihy1993transactional}, and our commit/rollback provenance (\code{transaction\_id}, \code{commit\_status}) and post-hoc cross-loop consistency check (L2, Section~\ref{sec:c3}) are directly in that tradition; the L2 check is a validation phase in the optimistic-concurrency sense. The novelty here is the \emph{unit} of speculation: a multi-cycle guidance transition of two coupled LLM agent loops, verified by re-executing the real miners and blinded evaluators, rather than tokens (speculative decoding) or memory words (transactional memory). A rejected draft costs speculative overhead but never corrupts the committed state, mirroring the safety property of speculative decoding while operating over a stateful, multi-agent transaction rather than a token sequence. The analogy is only partial. Speculative decoding gives an \emph{exactness} guarantee, in that the accelerated sampler reproduces the target model's output distribution. VST gives only a \emph{state-safety} guarantee: a rejected leap rolls back to an exact prior state, while an accepted leap is validated against empirical quality gates rather than proven distributionally identical to the unaccelerated trajectory. Quality preservation is therefore an empirically estimated bound (Equation~\eqref{eq:quality}), not a theorem.

%% file: sections/03_formulation.tex
\subsubsection{Dual-loop system substrate}
\label{sec:formulation-splits}
Before developing the structured transport that carries our contribution, we fix the substrate it wraps and the objective it must respect; nothing in this subsection is ours, and we state it only so the typing of C1 and its downstream leap have a precise object to attach to. We adopt, without modification, the dual-loop evolutionary search over formulaic alphas introduced in prior work \cite{alphasingularity}; we summarize it here only to fix notation, as it is the substrate on which our contributions operate rather than a contribution of this paper. A single \emph{ResearchReportAgent} advises both mining loops and receives structured feedback from both sides. In the \emph{alpha-discovery loop}, an \emph{AlphaMiner} proposes candidate factors and three independent \emph{AlphaEvaluator} instances produce diagnostic reports; their aggregate miner brief drives the next alpha cycle, and their report brief returns to the report agent. In the \emph{evaluation-optimization loop}, an \emph{EvaluationMiner} proposes metric-set and portfolio changes and three independent \emph{FactorMetricsEvaluator} instances analyze the resulting configuration and empirical evidence in the same multi-cycle pattern. The loops are coupled: the evaluation-side metric set scores alpha candidates, while accepted factors become evaluation-side samples.

To state the contribution boundary explicitly: the dual-loop topology, the six evaluators, and the metric/skill libraries that the evaluators evolve are inherited from prior work \cite{alphasingularity} and used unchanged. Everything this paper claims sits at and above the communication layer, and all of it rests on one load-bearing move: typing that layer (C1). The multi-horizon guidance predictor (C2), the transactional verify--leap controller (C3), and the \emph{separation} of the predictor's skill evolution from the inherited evaluator libraries (C4) are the three capabilities that typing makes possible, not independent layers. Where C4 discusses evaluator rubrics, it does so only to specify what the predictor must not influence, not to claim the evaluator-evolution mechanism itself.

Data are partitioned into four disjoint temporal splits: a training segment for alpha search, a validation segment for evaluator self-check and top-$K$ selection, a test segment, and a holdout segment. The test and holdout splits play different roles by design, and we keep both rather than collapsing them into one out-of-sample set. The test segment is visible to the meta-review agents: the evaluators may read test-segment outcomes when they diagnose configurations and feed guidance back into the loop, so the test split, while never used for training, is inside the system's optimization horizon and can carry a mild selection-induced leakage. The holdout segment is stronger: no agent, gate, horizon choice, or skill update is ever allowed to touch it, so it is the only partition that is free of every closed-loop decision. Reporting both lets a reader separate ``after the training period but still within the system's view'' (test) from ``fully isolated'' (holdout); an effect that survives only on the latter cannot be an artifact of the loop optimizing toward the evaluation segment. All model-selection signals are drawn from the training and validation partitions; the test and holdout partitions are isolated from horizon selection, skill updates, and every gate decision.

\subsubsection{Objective and quality constraint}
Let $S_t$ be the committed system state at feedback cycle $t$: the factor pool, evaluation configuration, report memory, and approved skills. Ordinary execution advances $S_t \to S_{t+1}$ by running one full miner--evaluator--report cycle. Our goal is to predict a future \emph{miner-directed guidance state}, use it to advance both loops by more than one cycle, validate the resulting actual outputs, and skip at least two ordinary cycles when validation succeeds. We report the cycle speedup
\begin{equation}
\mathrm{Speedup}_{\mathrm{cycle}} = T_{\mathrm{base}} / T_{\mathrm{leap}},
\end{equation}
and the corresponding wall-clock speedup, subject to a quality constraint on the committed leap state relative to the baseline optimum $\theta^\star$,
\begin{equation}
\mathrm{Quality}(\theta_{\mathrm{leap}}) \;\ge\; \mathrm{Quality}(\theta^\star) - \epsilon .
\label{eq:quality}
\end{equation}
Here $\theta^\star$ denotes the state the unaccelerated baseline reaches by running every ordinary feedback cycle to convergence under the same data splits, budget, and stopping rule; it is the reference the accelerated run must match within tolerance $\epsilon$, and it is computed only on development partitions. Equation~\eqref{eq:quality} is treated as an empirical constraint whose thresholds are fixed before final testing; if every trial fails, state evolution matches the baseline and incurs only the measured speculative overhead, which we report rather than claiming a cost-free worst case. With the substrate and objective fixed, the rest of the method follows from a single decision about how these agents communicate, which we take up next.

%% file: sections/04_c1_structured_transport.tex
\label{sec:c1}
The existing A2A implementation persists each round as a JSON object whose top-level namespaces are the operational payloads exchanged between agents; the agent logs retain the corresponding prompt--response trace. We do not replace this operational format. Instead we add a canonical routing wrapper around it, so that every inter-agent transmission becomes
\begin{equation}
m_{i\to j} = \big\langle\, e_{i\to j},\; x_{f(m)},\; p_{i\to j} \,\big\rangle \in \mathcal{S}_{r(i,j)},
\label{eq:wrapper}
\end{equation}
where $e$ is a universal envelope, $f(m)$ is the original A2A field path, $x_{f(m)}$ is that field's native string or dictionary payload, and $p$ is transactional provenance. For example, the field path \code{advise.to\_alpha\_miner} maps to the argument \code{rr\_advise} of the miner's \code{propose} call. The wrapper therefore separates routing and commit semantics from task content while retaining the receiver API used by the completed system.

\subsubsection{Design principles for A2A compatibility}
As anticipated in the Introduction, this protocol is the baseline system's own communication architecture made explicit, not a format invented to justify a leap. It rests on three principles for the record space $\mathcal{S}$. First, payloads are \emph{A2A-compatible}: canonicalization preserves the existing round-file field path and the receiving method argument, so structuring the channel changes no agent's task interface. Second, packets are \emph{causally addressable}: round, cycle, state, and source-artifact identifiers make every feedback transition replayable. Third, packets are \emph{unicast and visibility typed}: each physical packet has exactly one receiver, and a field is delivered only if its source segment and receiver role satisfy the access policy. Together these principles give the system one communication schema while preserving concrete producer, consumer, visibility, and state-transition semantics on every edge.

\subsubsection{Three-layer packet: envelope, body, and provenance}
The envelope $e$ contains four field groups, realized as a single typed record (\code{A2AEnvelope}):
\begin{itemize}
  \item \textbf{identity}: \code{message\_id}, \code{run\_id}, \code{round};
  \item \textbf{routing}: \code{sender}, \code{receiver}, \code{receiver\_arg};
  \item \textbf{causality}: \code{parent\_state\_id}, \code{cycle}, \code{sequence};
  \item \textbf{contract}: \code{payload\_path}, \code{schema\_version}, \code{visibility}.
\end{itemize}
The contract selects the schema for one native body $x_{f(m)}$. The identity fields, the parent-state link, and the sequence are generated deterministically by the gateway, because the existing round JSON stores payloads by namespace rather than as standalone messages. The provenance block $p$ links the packet back to its source round file and the producing agent's log, records whether the packet is \emph{observed} or \emph{speculative}, and records whether its transaction commits (\code{origin}, \code{transaction\_id}, \code{commit\_status}). Validation proceeds in a fixed order: validate the envelope, validate the body against the selected contract, then admit the packet to a transaction.

Table~\ref{tab:fields} separates the operational role of each layer from its later use by the predictor. The same fields that make ordinary transmission explicit also serve as stable trajectory coordinates, which is the mechanism by which C1 enables C2.

\begin{table}[t]
\caption{Field-by-field rationale for the structured communication plane. The left column is why a group helps ordinary transmission; the right column is why the same group helps later prediction.}
\label{tab:fields}
\centering
\footnotesize
\begin{tabular}{@{}p{0.21\linewidth}p{0.36\linewidth}p{0.36\linewidth}@{}}
\toprule
\textbf{Layer / field group} & \textbf{Helps transmission} & \textbf{Helps prediction} \\
\midrule
Envelope: identity & Deterministically identifies each transmission within a persisted round. & Records align into state-indexed trajectory observations $Z_t$. \\
Envelope: routing & Receiver and its method argument are explicit. & Unrelated edges stay distinct; inaccessible fields cannot leak into a target. \\
Envelope: causality & Receivers reconstruct ancestry; the controller can replay a feedback cycle. & The two loops are time-aligned and the future target $t{+}k$ is well defined. \\
Envelope: contract & Gateway validates the round-file field against its receiver schema and access policy. & Type identity fixes the decoder, loss, and aggregation rule per field. \\
Native body $x_{f(m)}$ & Retains operational keys (\code{rr\_advise}, \code{failure\_feedback}, \code{candidates}, \code{config}). & Prediction targets stay tied to the exact field paths and receiver arguments used at run time. \\
Provenance $p$ & Evidence is externally inspectable and traceable to its source artifact; speculative state stays isolated until commit. & Only committed prefixes supervise future bundles; rollback traces stay out of the state trajectory. \\
\bottomrule
\end{tabular}
\end{table}

\subsubsection{Deterministic unicast routing}
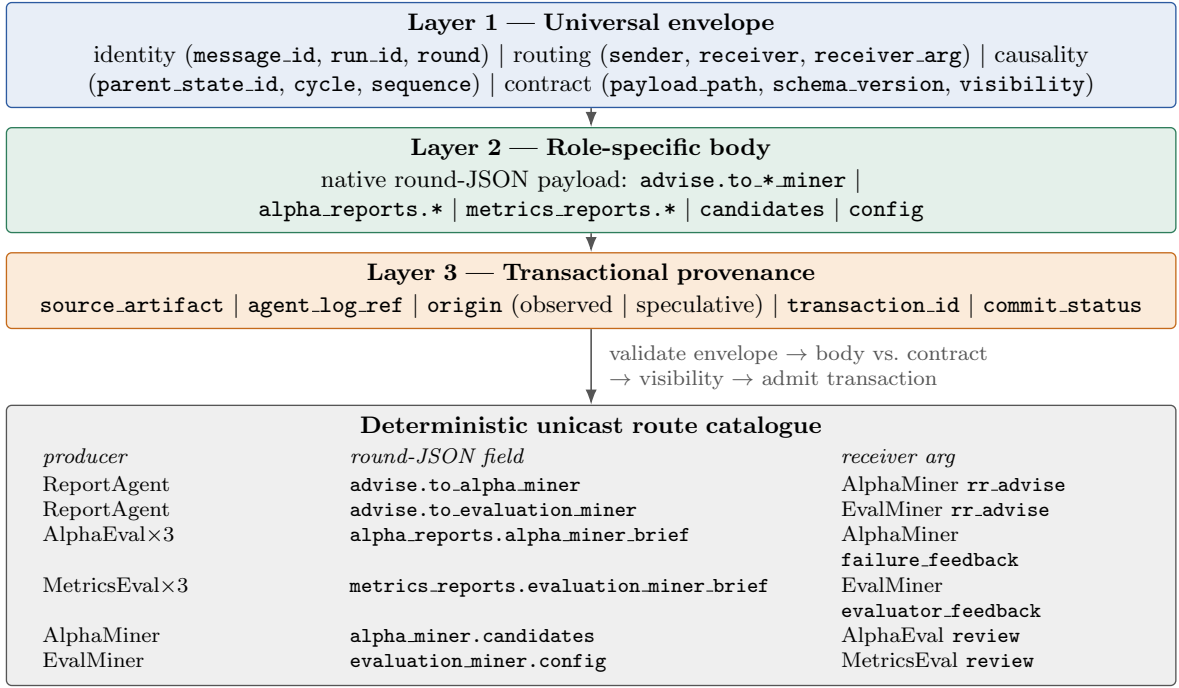
\begin{figure}[t]
\centering
\input{figures/fig_protocol}
\caption{A2A-compatible three-layer protocol and deterministic route catalogue. The wrapper adds identity, routing, causality, visibility, and provenance while retaining the original round-file \code{payload\_path} and receiver argument. Native round-JSON fields are wrapped, routed to existing method arguments, and retained as prediction targets.}
\label{fig:protocol}
\end{figure}

Routing is deterministic and unicast. The envelope names exactly one receiver; when a single logical producer result has content for multiple consumers, the gateway materializes one physical packet per destination. Such packets retain a shared parent identifier but may carry different consumer-specific body types. The role-specific contracts are:
{\sloppy
\begin{itemize}
  \item \textbf{AlphaMiner} receives two fields: \code{advise.to\_alpha\_miner} maps to \code{rr\_advise}, and \code{alpha\_reports.alpha\_miner\_brief} maps to \code{failure\_feedback}.
  \item \textbf{EvaluationMiner} receives \code{advise.to\_evaluation\_miner} as \code{rr\_advise} and \code{metrics\_reports.evaluation\_miner\_brief} as \code{evaluator\_feedback}.
  \item Each \textbf{AlphaEvaluator} receives \code{alpha\_miner.candidates} and an evaluation summary; the aggregate \code{alpha\_reports.research\_report\_brief} returns alpha-side findings to the report agent.
  \item Each \textbf{FactorMetricsEvaluator} receives the evaluation configuration, backtest samples, and the visible metric set; the aggregate \code{metrics\_reports.research\_report\_brief} reaches the report agent only when its visibility policy permits the edge.
\end{itemize}
\par}
Figure~\ref{fig:protocol} shows this hierarchy and the complete route catalogue.

\subsubsection{Visibility typing and leakage prevention}
The visibility field enforces an access policy that becomes essential once a predictor is added. Prediction packets are admitted only when routed to a miner, only when their field path is one of the four predicted paths, and only under \code{miners\_only} visibility; a packet addressed to an evaluator with a speculative origin is rejected before persistence. This rule is enforced by the gateway's validation step rather than by convention, so the six evaluators structurally cannot receive a forecast. Separating this policy into a typed field, rather than leaving it implicit in prose, is what lets the same channel safely carry both observed and speculative traffic.

\subsubsection{Worked example of a canonical packet}
The canonical wrapper around one actual advice transmission from the report agent to the AlphaMiner is a single record with an \code{envelope} block (identity, routing, causality, contract), a \code{body} carrying \code{rr\_advise}, and a \code{provenance} block naming the source round file, the producing agent log, an \code{origin} of \code{observed}, and a \code{commit\_status} of \code{committed}. All other edges use the same envelope with a different body schema.

Every edge now carries this structure. That is the whole of the load-bearing contribution; the three capabilities in the sections that follow, forecasting the trajectory, verifiably committing a leap over it, and co-evolving the predictor without corrupting its judges, are what it makes possible, and none of them could be built on a prose channel.

%% file: figures/fig_protocol.tex
\begin{tikzpicture}[vspfont, node distance=2.5mm]
\node[vspL1, text width=0.92\linewidth] (l1)
  {\textbf{Layer 1 --- Universal envelope}\\[1pt]
   identity (\texttt{message\_id}, \texttt{run\_id}, \texttt{round}) $\mid$
   routing (\texttt{sender}, \texttt{receiver}, \texttt{receiver\_arg}) $\mid$
   causality (\texttt{parent\_state\_id}, \texttt{cycle}, \texttt{sequence}) $\mid$
   contract (\texttt{payload\_path}, \texttt{schema\_version}, \texttt{visibility})};
\node[vspL2, text width=0.92\linewidth, below=of l1] (l2)
  {\textbf{Layer 2 --- Role-specific body}\\[1pt]
   native round-JSON payload: \texttt{advise.to\_*\_miner} $\mid$ \texttt{alpha\_reports.*} $\mid$
   \texttt{metrics\_reports.*} $\mid$ \texttt{candidates} $\mid$ \texttt{config}};
\node[vspL3, text width=0.92\linewidth, below=of l2] (l3)
  {\textbf{Layer 3 --- Transactional provenance}\\[1pt]
   \texttt{source\_artifact} $\mid$ \texttt{agent\_log\_ref} $\mid$
   \texttt{origin} (observed $\mid$ speculative) $\mid$ \texttt{transaction\_id} $\mid$ \texttt{commit\_status}};
\draw[vsparrow] (l1)--(l2);
\draw[vsparrow] (l2)--(l3);

\node[vspNeutral, text width=0.92\linewidth, below=10mm of l3] (routes)
 {\textbf{Deterministic unicast route catalogue}\\[2pt]
  {\scriptsize
  \begin{tabular}{@{}p{0.24\linewidth}p{0.40\linewidth}p{0.26\linewidth}@{}}
  \textit{producer} & \textit{round-JSON field} & \textit{receiver arg} \\[1pt]
  ReportAgent & \texttt{advise.to\_alpha\_miner} & AlphaMiner \texttt{rr\_advise} \\
  ReportAgent & \texttt{advise.to\_evaluation\_miner} & EvalMiner \texttt{rr\_advise} \\
  AlphaEval$\times$3 & \texttt{alpha\_reports.\allowbreak alpha\_miner\_brief} & AlphaMiner \texttt{failure\_feedback} \\
  MetricsEval$\times$3 & \texttt{metrics\_reports.\allowbreak evaluation\_miner\_brief} & EvalMiner \texttt{evaluator\_feedback} \\
  AlphaMiner & \texttt{alpha\_miner.candidates} & AlphaEval \texttt{review} \\
  EvalMiner & \texttt{evaluation\_miner.config} & MetricsEval \texttt{review} \\
  \end{tabular}}};
\draw[vsparrow] (l3.south) -- (routes.north)
  node[midway, right=1mm, text=vspGray, font=\scriptsize, align=left]
  {validate envelope $\rightarrow$ body vs.\ contract\\ $\rightarrow$ visibility $\rightarrow$ admit transaction};
\end{tikzpicture}

%% file: sections/05_c2_prediction.tex
\label{sec:c2}
Once C1 makes the channel a typed, replayable record, the committed hand-offs form a trajectory a model can be trained on.

\subsubsection{State-aligned communication trajectory}
Let $Z_t$ denote the committed, state-aligned communication graph at feedback cycle $t$. It contains every visibility-legal packet among the report agent, the two miners, both evaluator groups, and the controller: report advice, candidate and configuration packets, evaluator analyses, consumer-specific briefs, evidence references, and the resulting control event. The miner-directed guidance at cycle $t$ is extracted from four exact A2A field paths:
\begin{equation}
\begin{aligned}
A_t^{\alpha} &= Z_t[\code{advise.to\_alpha\_miner}], \\
\bar{R}_t^{\alpha} &= Z_t[\code{alpha\_reports.alpha\_miner\_brief}], \\
A_t^{\mathrm{eval}} &= Z_t[\code{advise.to\_evaluation\_miner}], \\
\bar{R}_t^{\mathrm{eval}} &= Z_t[\code{metrics\_reports.evaluation\_miner\_brief}], \\
G_t^{\alpha} &= (A_t^{\alpha}, \bar{R}_t^{\alpha}), \qquad
G_t^{\mathrm{eval}} = (A_t^{\mathrm{eval}}, \bar{R}_t^{\mathrm{eval}}).
\end{aligned}
\label{eq:paths}
\end{equation}
The two $A$ channels are the two strings emitted by the single report agent; the two $\bar{R}$ channels are the structured aggregates of the corresponding three evaluator reports. Future report advice is therefore predicted explicitly and delivered directly to the same miner argument as in ordinary execution.

\subsubsection{Multi-horizon prediction target}
For each candidate horizon $k \in \mathcal{K} = \{2, 4, \dots, K_{\max}\}$, the supervised target is the accumulated structured guidance that the two miners would receive over the next $k$ ordinary cycles,
\begin{equation}
B_{t,k} = \mathrm{Aggregate}\Big(\{ G_{t+\ell}^{\alpha}, G_{t+\ell}^{\mathrm{eval}} \}_{\ell=1}^{k}\Big),
\label{eq:target}
\end{equation}
where the type-aware aggregator composes set edits, resolves repeated promote/demote decisions by recency and support, and retains the latest bounded categorical adjustment. This is a future \emph{guidance} target, not a future factor, metric, verdict, or score.

The aggregator is type-directed rather than a single reducer: each of the four channels carries fields of known type, and \textsc{Aggregate} dispatches per field type (Algorithm~\ref{alg:agg}, Appendix~\ref{app:agg}). Set-valued fields (for example, the accepted metric set or the active factor tags) accumulate additions and removals in cycle order and collapse repeated promote/demote decisions on the same element to the most recent one weighted by how many cycles supported it. Numeric or bounded-categorical fields (for example, a weight or a horizon bias) keep the latest value within the schema's declared range. String advice fields keep the most recent committed text. The type-directed replay makes $B_{t,k}$ a deterministic function of the committed window, so the same prefix always yields the same target.

\subsubsection{Predictor architecture and type-aware losses}
The predictor reads a sliding window $Z_{t-w+1:t}$ and a requested $k$, then emits $(\widehat{B}_{t,k}, q_{t,k}, u_{t,k}) = P_\phi(Z_{t-w+1:t}, k)$: a schema-constrained bundle, a calibrated success confidence $q$, and an epistemic uncertainty $u$. Categorical, set-valued, and numeric fields use type-appropriate losses,
\begin{equation}
\begin{aligned}
\mathcal{L}_P ={}& \lambda_c\, \mathrm{CE}(\widehat{B}^{\mathrm{cat}}, B^{\mathrm{cat}}) + \lambda_s\, \mathrm{SetLoss}(\widehat{B}^{\mathrm{set}}, B^{\mathrm{set}}) \\
&+ \lambda_n \| \widehat{B}^{\mathrm{num}} - B^{\mathrm{num}} \|_1 + \lambda_{\mathcal{S}}\, \mathrm{Penalty}_{\mathcal{S}_B}(\widehat{B}) \\
&+ \lambda_q\, \mathrm{Brier}(q, y_{\mathrm{commit}}).
\end{aligned}
\label{eq:predloss}
\end{equation}
Here $\mathrm{CE}$ is cross-entropy on categorical fields, and $\mathrm{SetLoss}$ is a soft Jaccard (1 minus intersection-over-union) between the predicted and target element sets, which penalizes both missing and spurious set members without imposing an order. $\mathrm{Penalty}_{\mathcal{S}_B}$ is a schema-validity term that is zero when $\widehat{B}$ satisfies the bundle schema $\mathcal{S}_B$ and positive for each violated constraint (out-of-range numeric, unknown categorical level, illegal route), which trains the heads to emit admissible bundles. The final term is the Brier score of the calibrated confidence $q$ against the realized commit outcome $y_{\mathrm{commit}} \in \{0,1\}$. The weights $\lambda_c, \lambda_s, \lambda_n, \lambda_{\mathcal{S}}, \lambda_q$ balance the per-type terms and are selected on a development split; the schema-penalty weight $\lambda_{\mathcal{S}}$ is set high enough that admissibility dominates, since an inadmissible bundle is rejected at L1 regardless of its content.
The predictor is one component containing an open-weight LLM, a shared trajectory encoder, a trajectory-skill library, an RL-trained horizon policy, and four typed decoding heads,
\begin{equation}
\widehat{B}_{t,k} = \big( \widehat{A}_{t,k}^{\alpha},\, \widehat{\bar{R}}_{t,k}^{\alpha},\, \widehat{A}_{t,k}^{\mathrm{eval}},\, \widehat{\bar{R}}_{t,k}^{\mathrm{eval}} \big).
\label{eq:heads}
\end{equation}
In our implementation the open-weight LLM is \code{gpt-oss-120b}, served locally through an OpenAI-compatible endpoint; it is the only agent that receives a trainable local model, while the miner, evaluator, and report agents keep their existing hosted-API configuration. The base checkpoint is frozen and only a LoRA adapter is updated, first by supervised fine-tuning on causal trajectory prefixes and then by the constrained policy update of Section~\ref{sec:c4}.
The four channels share one global state representation because they are causally coupled. In this sense the predictor plays the role of a learned dynamics model over the communication trajectory, analogous to latent world models used for planning \cite{hafner2019planet,hafner2020dreamer}, except that its forecast is never trusted directly: it is validated by real execution (Section~\ref{sec:c3}). The communication layer serializes this logical action as two unicast packets, each addressed directly to its miner and consumed as one \code{propose} call:
\begin{equation}
\begin{aligned}
\hat{m}_{t,k}^{\alpha}:\;& \code{sender}{=}\code{predictor\_agent},\ \code{receiver}{=}\code{alpha\_miner}, \\
& \code{rr\_advise} \leftarrow \widehat{A}_{t,k}^{\alpha},\ \code{failure\_feedback} \leftarrow \widehat{\bar{R}}_{t,k}^{\alpha}, \\
\hat{m}_{t,k}^{\mathrm{eval}}:\;& \code{sender}{=}\code{predictor\_agent},\ \code{receiver}{=}\code{evaluation\_miner}, \\
& \code{rr\_advise} \leftarrow \widehat{A}_{t,k}^{\mathrm{eval}},\ \code{evaluator\_feedback} \leftarrow \widehat{\bar{R}}_{t,k}^{\mathrm{eval}}.
\end{aligned}
\label{eq:destpackets}
\end{equation}
Both have \code{origin=speculative} and the same parent state. No third packet is created for an evaluator or the report agent. Four predicted structures thus require four output heads, but neither four predictor agents nor four independent RL policies.

\subsubsection{Training corpus and no-leakage guarantee}
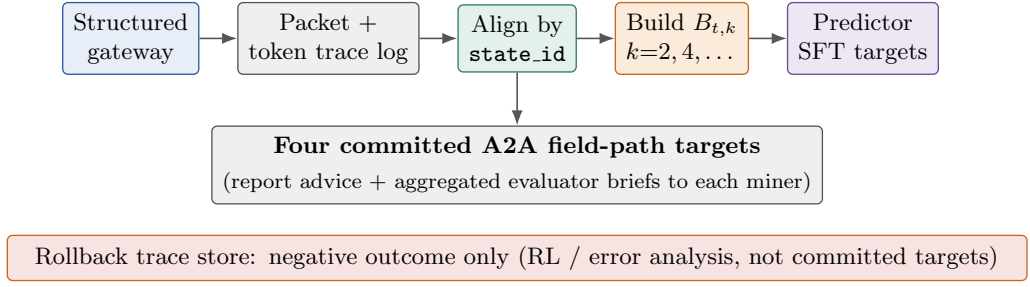
\begin{figure}[t]
\centering
\input{figures/fig_predpipe}
\caption{Predictor data pipeline. The structured gateway records committed packets and token-faithful model traces, aligns four miner-directed target channels by state, and composes them into two destination-specific guidance packets at each direct horizon. Rollback traces are stored separately for RL and error analysis, not inserted into the committed supervised trajectory.}
\label{fig:predpipe}
\end{figure}

The structured gateway already validates every packet, so it also appends the committed packet, prompt, sampled tokens, log probabilities, and state identifier to an immutable trajectory log. Sampled assistant tokens are copied verbatim; only fixed scaffolding is retokenized, and a loss mask marks model tokens as trainable. ``Deterministic replay'' has a specific meaning here: because the producing agents are LLMs (hosted-API miners and evaluators, and a local predictor model) whose sampling is not reproducible by re-invocation, replay is the deterministic reconstruction of a trajectory \emph{from the recorded log of tokens and states}, not deterministic re-execution of the agents. Likewise, ``immutable'' refers to an append-only log that is never edited in place after commit; we do not claim tamper-evidence or content-addressed hashing beyond append-only discipline. The target constructor projects the committed graph onto the four state-aligned paths of Equation~\eqref{eq:paths}. Training examples pair the same global causal prefix with several direct horizons, so a horizon-eight target is learned from the observed window $t{+}1{:}t{+}8$ rather than generated by chaining eight one-step predictions. This is the direct (as opposed to recursive) multi-step forecasting strategy \cite{chevillon2007direct}, which avoids compounding one-step error across the horizon; what is specific to our setting is that the predicted object is a typed structured guidance bundle over a communication graph rather than a scalar or vector series. At cycle $t$, the prefix contains only packets committed by $t$; controller-only evidence is excluded by the visibility field; the test and holdout partitions never form guidance targets, select a horizon, update a skill, or compute a reward; and rolled-back traces are labeled failures and cannot become positive supervised targets. Figure~\ref{fig:predpipe} summarizes the pipeline.

%% file: figures/fig_predpipe.tex
\begin{tikzpicture}[vspfont, node distance=5mm]
\node[vspL1] (gw) {Structured\\ gateway};
\node[vspNeutral, right=of gw] (log) {Packet +\\ token trace log};
\node[vspL2, right=of log] (align) {Align by\\ \texttt{state\_id}};
\node[vspL3, right=of align] (build) {Build $B_{t,k}$\\ $k{=}2,4,\dots$};
\node[vspL4, right=of build] (sft) {Predictor\\ SFT targets};
\draw[vsparrow] (gw)--(log); \draw[vsparrow] (log)--(align);
\draw[vsparrow] (align)--(build); \draw[vsparrow] (build)--(sft);
\node[vspNeutral, align=center, font=\footnotesize, below=7mm of align] (four)
 {\textbf{Four committed A2A field-path targets}\\[2pt]
  {\scriptsize (report advice + aggregated evaluator briefs to each miner)}};
\node[vspReject, text width=0.8\linewidth, below=4mm of four] (rb)
 {Rollback trace store: negative outcome only (RL / error analysis, not committed targets)};
\draw[vsparrow] (align.south) -- (four.north);
\end{tikzpicture}

%% file: sections/06_c3_verify_leap.tex
\label{sec:c3}
A forecast is only useful if an incorrect one is inexpensive to reverse. Because C1 records provenance and commit status per packet and isolates speculative traffic until it commits, the leap can be run as a transaction: propose the forecast, execute it for real in a scratch state, and either commit the whole transition or discard it with no trace left in the committed trajectory.

Given a predicted bundle $\widehat{B}_{t,k}$, the two miners produce actual outputs $O_t = (O_t^{\alpha}, O_t^{\mathrm{eval}})$ in a scratch namespace. The six evaluators then produce independent reports $E_t = (E_{t,1:3}^{\alpha}, E_{t,1:3}^{\mathrm{metrics}})$. The controller validates this \emph{speculative outcome}, not whether the miner output matches the predicted guidance field by field. Four levels short-circuit in ascending order of cost.

\subsubsection{Adaptive horizon selection}
At state $S_t$ the predictor scores every candidate horizon and the controller chooses the farthest admissible one,
\begin{equation}
k_t = \max\big\{ k \in \mathcal{K} : k \ge 2,\ q_{t,k} \ge \tau_q,\ u_{t,k} \le \tau_u,\ \widehat{G}_{t,k} > 0 \big\},
\label{eq:horizon}
\end{equation}
where $\widehat{G}_{t,k}$ is the predicted saved compute after speculative and verification costs. If the set is empty, the system runs one normal feedback cycle and makes no leap. For an admissible $k_t$, the controller opens a scratch transaction and routes the two payload branches only to their respective miners; the scratch execution may use up to $b$ coupled refinements to resolve cross-loop inconsistency, but a trial counts as a speedup only when $b < k_t$.

\subsubsection{Four-level validation gate}
\paragraph{L1: Bundle legality and route policy}
\begin{equation}
V_1 = \mathbf{1}\big[ \widehat{B}_{t,k} \models \mathcal{S}_B \ \wedge\ \mathrm{targets}(\widehat{B}_{t,k}) = \{\code{alpha\_miner}, \code{evaluation\_miner}\} \big].
\label{eq:l1}
\end{equation}
This rejects malformed fields, stale parents, $k < 2$, and any evaluator-targeted prediction before speculative execution.

\paragraph{L2: Actual-output validity and cross-loop consistency}
\begin{equation}
V_2 = \mathbf{1}\big[ O_t^{\alpha} \models \mathcal{S}_\alpha \wedge O_t^{\mathrm{eval}} \models \mathcal{S}_{\mathrm{eval}} \wedge \mathrm{CrossConsistent}(O_t^{\alpha}, O_t^{\mathrm{eval}}, S_t) \big].
\label{eq:l2}
\end{equation}
The factor batch must be executable, the evaluation configuration legal, and the metric-set, sample, portfolio, and state references must agree. The gate does not require either output to equal a value in the predicted bundle. $\mathrm{CrossConsistent}(O_t^{\alpha}, O_t^{\mathrm{eval}}, S_t)$ returns true only when three referential checks hold: (i) every factor the evaluation side scores exists in the alpha side's produced batch (no dangling reference); (ii) the metric set named by $O_t^{\mathrm{eval}}$ is a subset of the metrics admissible at state $S_t$; and (iii) the portfolio in $O_t^{\mathrm{eval}}$ is built only from factors that pass the alpha side's own executability check. The check is an optimistic-concurrency validation phase in the sense of validation-based concurrency control: the two coupled loops execute speculatively and their joint state is checked for conflicts before commit. These three referential checks are necessary conditions we found sufficient to catch the divergences observed in our runs, not a proven-complete characterization of cross-loop consistency; a conflict outside this enumerated set would be admitted, so the check is a conservative screen rather than a soundness guarantee. Any violation fails L2, because it means the two loops have diverged into an incoherent joint state that no downstream quality score would be meaningful on.

\paragraph{L3: Blinded evaluator risk}
Each three-instance group writes reports independently before aggregation. The controller maps schema-defined critical findings to robust risk scores while preserving report-level severity and disagreement,
\begin{equation}
V_3 = \mathbf{1}\big[ \mathrm{RiskAgg}(E_{t,1:3}^{\alpha}) \le \tau_\alpha \ \wedge\ \mathrm{RiskAgg}(E_{t,1:3}^{\mathrm{metrics}}) \le \tau_m \big].
\label{eq:l3}
\end{equation}
$\mathrm{RiskAgg}$ maps each report's schema-tagged findings to a bounded severity score (a critical finding dominates a major, which dominates a minor) and then aggregates the three independent scores with a disagreement-preserving statistic: it takes the worst-case severity rather than the mean, so a single instance flagging a critical issue is not averaged away by two lenient ones. L3 is thus conservative by construction and keeps the three evaluators' independence meaningful.

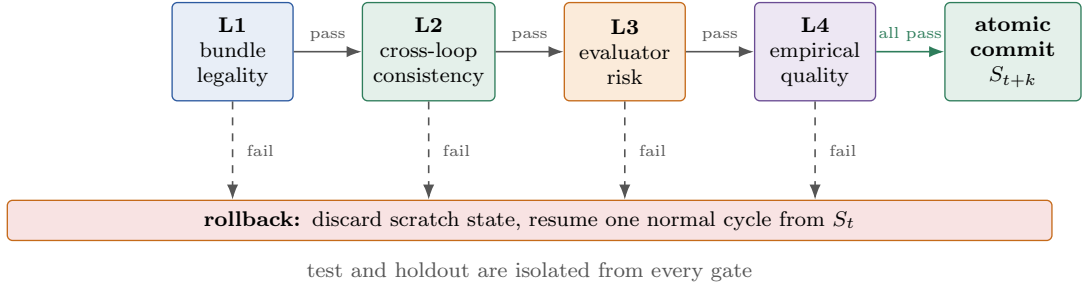
\begin{figure}[t]
\centering
\input{figures/fig_gate}
\caption{Four-level validation of a speculative outcome. The bundle is checked first; the actual miner outputs, blinded evaluator reports, and development-set evidence are checked next. Only an all-pass path atomically commits; any failure discards the scratch transaction and resumes one normal cycle from $S_t$.}
\label{fig:gate}
\end{figure}

\paragraph{L4: Empirical development gate}
The controller recomputes factor and portfolio quality from the miners' actual outputs on two development partitions, a train gate and an internal leap-validation split,
\begin{equation}
V_4 = \mathbf{1}\Big[ \min_{d} Q_{\alpha,d}(O_t^{\alpha}) \ge \tau_g \ \wedge\ \min_{d} Q_{m,d}(O_t) \ge \tau_p \Big],
\label{eq:l4}
\end{equation}
with $d \in \{\text{train-gate}, \text{leap-valid}\}$, where $Q_{\alpha,d}$ combines IC/RankIC stability and $Q_{m,d}$ combines portfolio return, risk, monotonicity, and cost. Final test and holdout data are isolated from all adaptive decisions. The transaction decision is
\begin{equation}
C_t = \prod_{\ell=1}^{4} V_\ell, \qquad C_t = 1 \Rightarrow \text{AtomicCommit}, \quad C_t = 0 \Rightarrow \text{Rollback}.
\label{eq:decision}
\end{equation}
Figure~\ref{fig:gate} shows the short-circuit structure.

\subsubsection{Transactional commit and rollback semantics}
We use transactional vocabulary (\code{transaction\_id}, \code{commit\_status},
atomic commit, rollback) in a narrowly restricted sense, and state the model
explicitly to avoid overclaiming the ACID properties the terms carry in a database.
The concurrency model is \emph{single-writer and strictly serial}: exactly one
speculative transaction is in flight at a time, rooted at the current committed
state, and the controller is its sole coordinator. Because execution is strictly
serial, serializability holds trivially: there is no concurrent schedule to reorder,
so isolation is immediate and we make no claim about a concurrency-control protocol.
``Atomic commit''
means the validated state is installed only after all four gates pass and the commit
is applied as a single controller step from the workflow's point of view; we do
not claim crash-atomicity or durability (a process failure during the
multi-store install is out of scope, not recovered by a write-ahead log), and
``exact rollback'' is best-effort restoration of the application-level committed
stores (factor pool, report memory, evaluator and predictor libraries) to $S_t$,
established by construction over the state we track rather than proved against
arbitrary side effects. Under this model a speculative trial is a transaction rooted
at \code{parent\_state\_id}. Its namespace contains both miner outputs, all six evaluator reports, the report agent's transaction-local synthesis, empirical artifacts, and any proposed changes to the predictor skill library or the evaluator libraries. None is visible in the committed databases or active agent memories before $C_t = 1$. On commit, the controller installs the validated factor and configuration state, sanitized report feedback, and approved skill updates as a single logical step (absent process failure, per the scope above). On rollback, it deletes the entire mutable namespace and resumes one ordinary feedback cycle from $S_t$; the immutable audit record is retained with \code{commit\_status=rollback} for RL and failure analysis but is excluded from the active factor pool, report memory, evaluator libraries, and positive supervised targets. This preserves an exact state rollback while still learning from failed speculation. A failed trial adds runtime overhead; it does not alter the baseline state transition.

\subsubsection{State safety and quality control}
Rollback gives deterministic state safety, while the quality of a committed leap depends on gate calibration. We therefore estimate
\begin{equation}
\Pr\big[\, \mathrm{Quality}(S_{t+k}) < \mathrm{Quality}_{\mathrm{base}}(S_t, k) - \epsilon \ \big|\ C_t = 1 \,\big] \le \delta
\label{eq:safety}
\end{equation}
on development runs, with thresholds fixed before final testing. The selected $k$ is a direct target, not a chain of one-step forecasts; a successful horizon-$k$ trial commits the validated scratch result as $S_{t+k}$ and records that $k$ ordinary guidance cycles were replaced by $b$ speculative refinements. The next decision is always anchored at the latest committed state.

%% file: figures/fig_gate.tex
\begin{tikzpicture}[vspfont, node distance=9mm,
    gate/.style={minimum width=16mm, minimum height=13mm, align=center, font=\scriptsize}]
\node[vspL1, gate] (l1) {\textbf{L1}\\ bundle\\ legality};
\node[vspL2, gate, right=of l1] (l2) {\textbf{L2}\\ cross-loop\\ consistency};
\node[vspL3, gate, right=of l2] (l3) {\textbf{L3}\\ evaluator\\ risk};
\node[vspL4, gate, right=of l3] (l4) {\textbf{L4}\\ empirical\\ quality};
\draw[vsparrow] (l1)--node[midway,above,font=\tiny,text=vspGray]{pass}(l2);
\draw[vsparrow] (l2)--node[midway,above,font=\tiny,text=vspGray]{pass}(l3);
\draw[vsparrow] (l3)--node[midway,above,font=\tiny,text=vspGray]{pass}(l4);

\node[vspbox, fill=vspGreenBg, draw=vspGreen, align=center, font=\scriptsize,
      minimum width=18mm, minimum height=13mm, right=of l4] (commit)
      {\textbf{atomic}\\ \textbf{commit}\\ $S_{t+k}$};
\draw[vsparrow, draw=vspGreen] (l4)--node[midway,above,font=\tiny,text=vspGreen]{all pass}(commit);

\coordinate (mid) at ($(l2.south east)!0.5!(l3.south west)$);
\node[vspReject, align=center, font=\scriptsize, text width=0.82\linewidth,
      below=13mm of mid, anchor=north] (rb)
      {\textbf{rollback:} discard scratch state, resume one normal cycle from $S_t$};
\foreach \n in {l1,l2,l3,l4}{%
  \draw[vspdash] (\n.south) -- (\n.south |- rb.north)
    node[midway,right=0.5mm,font=\tiny,text=vspGray]{fail};}
\node[below=1.5mm of rb, text=vspGray, font=\scriptsize]
      {test and holdout are isolated from every gate};
\end{tikzpicture}

%% file: sections/07_c4_coevolution.tex
\label{sec:c4}
The predictor improves by learning from outcomes, and it learns safely because the evaluators that judge it never see what it predicted. Both are consequences of C1: the commit/rollback verdict is a typed, replayable outcome the policy can be trained on, and the same visibility typing that keeps forecasts out of evaluator prompts keeps their skill libraries structurally separate.

The predictor is not frozen after supervised training. Its skill library stores reusable procedures for type-aware feedback aggregation, cross-loop consistency, trajectory change detection, horizon calibration, and rollback diagnosis. The RL policy selects and composes these skills when producing $(k_t, \widehat{B}_{t,k})$, and commit or rollback outcomes then provide outcome-grounded supervision as the mining regime changes.

\subsubsection{Agentic reinforcement learning of the predictor}
The committed trajectories support online policy improvement inside the predictor. Whereas single-agent LLM policies are typically trained to produce an answer or an action sequence \cite{yao2023react,shinn2023reflexion} and optimized with policy-gradient methods such as PPO \cite{schulman2017ppo} or its group-relative variant \cite{shao2024deepseekmath}, our predictor is optimized against a \emph{transaction-level} outcome from a coupled dual loop. The RL state is $Z_{t-w+1:t}$; the joint action is $a_t = (k_t, \widehat{B}_{t,k_t})$, where $\widehat{B}$ contains all four typed channels; and the environment is the transactional dual loop consisting of both miners, six blinded evaluators, the report interaction, and the controller. We use one predictor with a shared policy and four typed heads (rationale below). During supervised initialization each head has a schema-appropriate auxiliary loss; during online RL the complete four-channel action receives one transaction-level reward. Let $C_t \in \{0,1\}$ be the controller decision, $C_{\mathrm{saved}}(k_t, b)$ the measured cost of the skipped ordinary cycles, and $C_{\mathrm{trial}}$ the speculative plus verification cost. We use
\begin{equation}
r_t = C_t\big[ C_{\mathrm{saved}}(k_t, b) - C_{\mathrm{trial}} \big] - (1 - C_t) C_{\mathrm{trial}} - \lambda_r R_t - \lambda_q D_t,
\label{eq:reward}
\end{equation}
where $R_t$ is evaluator risk and $D_t$ is any development-set quality deficit. This rewards a long leap only when it both commits and saves real work; rollback traces train the policy to shorten the horizon or change guidance in similar states. Supervised learning fits the bundle content, RL fits horizon and utility, and the skill library supplies reusable reasoning procedures; all three belong to the same predictor.

We claim no new optimizer here; the contribution is two design choices the structured, transactional channel makes available. One is representational: a single shared-encoder policy emits all four typed heads, rather than four RL agents, because the channels are one coupled state and splitting them would turn cross-channel consistency into a non-stationary coordination problem. The other concerns where the reward comes from: the four-level gate serves as the RL environment, so the learning signal is its commit/rollback verdict on the real dual loop rather than a proxy score, and the trajectories it rejects supply the policy's negatives. Neither choice is available on a prose channel, which offers no typed joint action to share and no atomic outcome to reward against.

\subsubsection{Skill lifecycle}
Predictor skills evolve by three transactional operations, echoing the growing executable-skill library of open-ended agents \cite{wang2023voyager} but gated by transactional outcomes rather than task success alone. \emph{Acquisition}: a recurring feedback-transition pattern with high horizon-conditioned bundle loss can spawn a specialized trajectory skill. \emph{Reinforcement}: a skill gains selection weight only when its bundle causes the actual dual loop to pass L1--L4 and produce positive net cycle savings. \emph{Pruning or repair}: rollback reason, evaluator risk, and field-level bundle residuals identify whether to shorten a skill's horizon, constrain its output, or rewrite it. Updates occur after the outcome is known, never during scoring.

\subsubsection{Separation of predictor and evaluator skill libraries}
The six evaluators do not use predicted guidance. Each AlphaEvaluator receives only the actual candidate batch, its development-side evaluation summary, the allowed metric dimensions, and its assigned historical view; it writes an independent report and may propose a new or revised rubric. Each FactorMetricsEvaluator receives only actual factor and backtest samples, portfolio output, the metric set, and its assigned historical view; it may propose a meta-rubric. The three reports on each side are aggregated into distinct miner and report-agent briefs while preserving report-level severity and disagreement. Evaluator proposals are transaction-local while the controller is deciding: if the leap commits, proposals supported by the verified evidence enter the normal trial-to-promotion process; if it rolls back, the proposals are discarded with the scratch state, and the labeled actual-output trace remains available as a negative example for later rubric repair. Predictor identifiers, $k$, confidence, uncertainty, and predicted fields are absent from every evaluator prompt, which prevents forecast anchoring.

\begin{figure}[t]
\centering
\input{figures/fig_coevo}
\caption{Three transactional evolution paths. Predictor skills learn from commit/rollback outcomes. Evaluators evolve rubric and meta-rubric libraries from actual outputs only, never seeing the prediction, and their updates install only through the post-decision promotion path.}
\label{fig:coevo}
\end{figure}
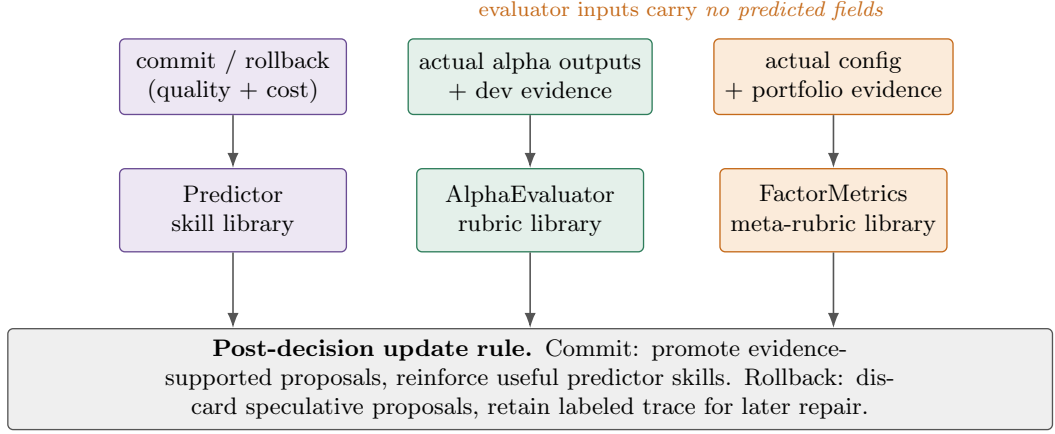

Keeping these update paths separate is a safety property with a specific failure it prevents. Predictor skills optimize forecasting and horizon choice; evaluator rubrics optimize diagnosis of actual outputs. If a single successful predictor could also shape the evaluator rubrics, it could gradually teach the evaluators to agree with its own forecasts, and the L3 gate would lose its independence. This is the reward-hacking hazard of a coupled learner and its own judge \cite{amodei2016concrete}: an optimizer with influence over its evaluation criterion tends to game the criterion rather than the underlying objective. The transactional promotion rule additionally keeps speculative failures from silently changing active evaluation standards. Figure~\ref{fig:coevo} shows the three separate evolution paths.

%% file: figures/fig_coevo.tex
\begin{tikzpicture}[vspfont, node distance=7mm]
\node[vspL4, minimum width=30mm, minimum height=10mm] (o1) {commit / rollback\\ (quality + cost)};
\node[vspL2, minimum width=30mm, minimum height=10mm, right=8mm of o1] (o2) {actual alpha outputs\\ + dev evidence};
\node[vspL3, minimum width=30mm, minimum height=10mm, right=8mm of o2] (o3) {actual config\\ + portfolio evidence};

\node[vspL4, minimum width=30mm, minimum height=11mm, below=of o1] (p) {Predictor\\ skill library};
\node[vspL2, minimum width=30mm, minimum height=11mm, below=of o2] (a) {AlphaEvaluator\\ rubric library};
\node[vspL3, minimum width=30mm, minimum height=11mm, below=of o3] (m) {FactorMetrics\\ meta-rubric library};

\draw[vsparrow] (o1)--(p);
\draw[vsparrow] (o2)--(a);
\draw[vsparrow] (o3)--(m);
\node[text=vspOrange, font=\scriptsize, align=center, above=1mm of $(o2.north)!0.5!(o3.north)$]
  {evaluator inputs carry \emph{no predicted fields}};

\node[vspNeutral, text width=0.82\linewidth, align=center, below=10mm of a, anchor=north] (rule)
 {\textbf{Post-decision update rule.}
  Commit: promote evidence-supported proposals, reinforce useful predictor skills.
  Rollback: discard speculative proposals, retain labeled trace for later repair.};
\foreach \n in {p,a,m}{\draw[vsparrow] (\n.south) -- (\n.south |- rule.north);}
\end{tikzpicture}

%% file: sections/08_experiments.tex

Our claims are of two kinds, and the experiments separate them. The comparison
against baselines (Sections~\ref{sec:exp-holdout}--\ref{sec:exp-multimetric}) asks
whether the verified structured search yields factors that survive out of sample; the
transport ablation (Section~\ref{sec:exp-ablation}) asks the load-bearing question
directly, whether the typing of C1 earns its place through predictive accuracy or, as
we argue, through the correctness properties that C2--C4 depend on. The second
question is the one that decides whether structure, rather than speed or returns, is
the contribution, and we keep it in view throughout.

\subsection{Setup: Universe, Splits, and Baselines}
\label{sec:exp-setup}
We evaluate on the CSI~1000 (ZZ1000) constituent universe with disjoint temporal
splits for training, validation, test, and holdout; the holdout partition is never
touched by the closed loop and is used only for final out-of-sample reporting.
Every method mines factors on the training split and is filtered to valid
candidates on the validation split under a common gate (Sharpe~$\ge 1$ and
annualized excess return~$\ge 0.05$); no top-$K$ post-selection is applied. All
reported quality figures are computed with a single shared 10-group backtest
evaluator so that cross-method numbers are directly comparable. All return and
risk-adjusted figures reported in this paper are \emph{gross}, before transaction
costs, slippage, and turnover constraints; because every method is scored under the
identical shared backtest, the comparison is internally consistent, but the absolute
magnitudes are not net-of-cost and should not be read as deployable returns. On a
mid- and small-cap universe such as CSI~1000, turnover and cost adjustment can be
material; we treat a net-of-cost evaluation as out of scope for this study and note
it as a limitation (Section~\ref{sec:limitations}).

We compare against seven baselines spanning hand-crafted, symbolic, reinforcement-learning,
and LLM-agent alpha discovery: Alpha101 (B1) \cite{kakushadze2016alpha101}, a genetic-programming
searcher (B2, the gplearn baseline from the AlphaGen repository \cite{yu2023alphagen}), DSO/DSR (B3) \cite{petersen2021dsr},
AlphaGen (B4) \cite{yu2023alphagen}, AlphaMCTS (B5) \cite{shi2025alphamcts}, AlphaAgent (B6)
\cite{tang2025alphaagent}, and an automated Alpha-GPT approximation (B7) \cite{wang2025alphagpt}.
Our system is denoted OURS (M4). We report, per method and per split,
the number of valid factors $n_{\mathrm{valid}}$, the fraction of factors with backtest
Sharpe above~1 and above~2, and the median Sharpe. To compare per-factor Sharpe
distributions between OURS and each baseline we use the Mann--Whitney $U$ statistic,
which we summarize as the \emph{probability of superiority}
$\hat{P}(\text{OURS} > \text{baseline}) = U / (n_1 n_2)$ (the probability that a
uniformly drawn OURS factor has larger Sharpe than a uniformly drawn baseline
factor). We report this rank effect size rather than a tail $p$-value:
because factors produced within one run share a training window and the same
closed-loop selection, they are not independent, so the effective sample size is far
below $n_{\mathrm{valid}}$ and an i.i.d.\ $p$-value would be misleadingly small
(orders of magnitude too significant). The probability of superiority is a
descriptive rank statistic on a single committed pool per method that does not
presuppose independence; it is not a test over independent runs, and between-run
variance is not bounded here (see Section~\ref{sec:limitations}).

The valid-factor counts are matched across methods by budget rather than left to
differ by run length. Each method contributes the valid factors it accumulates in
mining order up to a common cap of $500$; the baselines emit a one-shot batch at
their configured budget (here $\le 500$, with B4 AlphaGen reaching only $200$ and B1
providing the $14$ hand-crafted Alpha101 formulas), and for our continual system we
take the first $500$ valid factors of the run in mining order, matching the
baselines' ``accumulate until the budget is reached'' rule rather than selecting the
Sharpe-best $500$. Because the per-factor quality metrics (median Sharpe,
$P(\mathrm{sh}{>}1)$) and the rank-based probability of superiority are computed per
factor and are not normalized by count, this cap removes any concern that a larger
$n$ inflates them; we report $n$ explicitly so the reader can weigh each comparison.
We are explicit about one property of this particular matching rule: for our
continual system the first $500$ valid factors are not a representative sample of the
whole run. That early segment has a higher holdout median Sharpe ($+0.34$) than the
run as a whole ($+0.13$ over all $2300$ scored factors) or its later segments
($\approx +0.10$), so ``first in mining order'' happens to select the stronger part
of our pool, whereas each baseline contributes its entire one-shot batch. The
matched-budget median in Table~\ref{tab:holdout} should therefore be read together
with the full-pool median ($+0.13$), and the load-bearing budget-controlled
comparison is the \emph{order-agnostic} top-$K$-by-validation-Sharpe result of
Section~\ref{sec:exp-topk}, which ranks each method by an identical rule and does not
depend on mining order; our out-of-sample edge persists there at $K=100$ and $K=200$.
The baselines were run under their own recommended search budgets and configurations
rather than weakened: that they reach strong in-sample fit supports this, since on
the validation split several sit between $1.5$ and $1.8$ median Sharpe (for example
B5 AlphaMCTS at $1.838$; Table~\ref{tab:splits}). Their out-of-sample collapse
is therefore a robustness gap, not an artifact of under-tuning.

All data splits, the common validity gate, and every gate threshold are defined by a
fixed experiment configuration (\code{config\_formal\_v1.yaml}) committed before final
testing; the test and holdout partitions are read only after these values are frozen,
so no threshold or horizon choice is informed by out-of-sample data.

\subsection{Out-of-sample comparison}
\label{sec:exp-holdout}
Table~\ref{tab:holdout} reports the holdout split, the strictest out-of-sample test.
What matters most for a trading application is realized return, which we report first:
on holdout, our system is the only method with a positive median annualized
return ($+0.059$); every baseline loses money out of sample, including the strongest
learned ones (B4 AlphaGen $-0.001$, B5 AlphaMCTS $-0.015$, B7 Alpha-GPT $-0.031$).
The same ordering holds on a risk-adjusted basis: our system attains a median Sharpe
of $0.346$, again the only positive median, while every baseline degrades into
negative territory (B4 $-0.006$, B5 $-0.056$, B7 $-0.159$).
OURS also attains the highest fraction of high-quality factors, with
$P(\mathrm{Sharpe}{>}1)=0.313$ (versus $0.224$ for the next-best baseline, Table~\ref{tab:holdout}) and, on the stricter $P(\mathrm{Sharpe}{>}2)=0.114$, again the highest of any method. The rank effect size points the
same way: the probability that a uniformly drawn OURS factor has higher holdout
Sharpe than a uniformly drawn baseline factor is above $0.5$ against every baseline,
though the margin is small for the strongest learned baselines and only large
against the fully collapsed ones ($\hat{P}(\text{OURS}{>}\text{base})=0.56$ vs.\ B4,
$0.58$ vs.\ B5, $0.57$ vs.\ B6, $0.57$ vs.\ B7, $0.66$ vs.\ B2, and $0.87$ vs.\ B3;
$0.64$ vs.\ B1 on only $14$ factors). We report these as descriptive rank statistics,
not powered tests: per-factor units within a run are not independent, so we do not
attach $p$-values (Section~\ref{sec:exp-setup}). The measured result is that OURS
holds a consistent but small per-factor rank advantage over the competitive learned
baselines (B4--B7) and a large one only where the baseline has collapsed out of
sample.

The positive figure is an
\emph{absolute} annualized return; the median \emph{excess} return over the CSI~1000
benchmark remains negative for every method, including ours ($-0.028$), though ours
is the smallest shortfall. The precise statement is therefore that our system is the
only one to produce a positive absolute out-of-sample return and the one whose excess
return decays least, not that it beats the benchmark on holdout.

The $0.346$ median is moderate in absolute terms; the claim is comparative. Under the shared backtest, ours is the only method that stays positive out of sample, partly because the learned baselines overfit into negative territory rather than because our factors are strong in absolute terms.

{\sloppy
To quantify the sampling uncertainty of these medians we compute a
$95\%$ bootstrap confidence interval by resampling each method's holdout factor
pool ($2000$ resamples, fixed seed). Our system's interval is
$[+0.207, +0.508]$, entirely above zero, whereas no baseline has a positive lower
bound: several sit entirely below zero (B2 $[-0.644,-0.524]$, B3
$[-1.706,-1.648]$) and the strongest learned baselines straddle zero (B4
$[-0.317,+0.127]$, B5 $[-0.296,+0.062]$). The positive-median result for our system
is therefore not an artifact of pool sampling. This
interval bounds the sampling variability of the median
within a single committed factor pool; it does not bound variability
across independent training runs, which would require repeated end-to-end
runs and remains a limitation (Section~\ref{sec:limitations}). For the same reason,
that our interval and a baseline's interval do not overlap is not a
between-method significance test: both intervals hold their respective single runs
fixed and exclude the dominant run-to-run variance component, so non-overlap shows
only that each single pool's median is stably estimated, not that the two underlying
search processes differ.\par}

\begin{table}[t]
\caption{Holdout comparison (CSI~1000, shared 10-group backtest; each method mined on train, filtered to valid on validation; no top-$K$ post-selection). $n_{\mathrm{valid}}$ is the validation-filtered pool size under a matched budget: every method contributes the valid factors it accumulates in mining order up to a common cap of $500$ (B4 AlphaGen reaches only $200$ and B1 provides $14$ hand-crafted formulas, reported honestly). For OURS this is the first $500$ valid factors of the continual run in mining order, not a Sharpe-ranked selection; $499$ of them produced a scored holdout backtest (one dropped for missing holdout coverage), and the holdout statistics are computed over those $499$. This early segment is stronger than the full run (full-pool median Sharpe $+0.13$; see Section~\ref{sec:exp-setup}), so this row should be read with the order-agnostic top-$K$ comparison of Table~\ref{tab:topk}. $^*$B7 is an automated Alpha-GPT approximation. The 95\% CI is a bootstrap interval for the median Sharpe ($2000$ resamples of each holdout pool); only our system's interval lies entirely above zero, which bounds within-pool median stability and is not a between-method test (Section~\ref{sec:exp-holdout}). The per-factor rank effect size $\hat{P}(\text{OURS}{>}\text{baseline})$ is reported in text; per-factor units within a run are not independent, so we give a descriptive rank statistic rather than a $p$-value (Section~\ref{sec:exp-setup}).}
\label{tab:holdout}
\centering
\footnotesize
\begin{tabular}{@{}lrrrrl@{}}
\toprule
Method & $n_{\mathrm{valid}}$ & $P(\mathrm{sh}{>}1)$ & med.\ AnnRet & med.\ sh & 95\% CI (sh) \\
\midrule
B1 Alpha101~\cite{kakushadze2016alpha101}  & 14   & 0.000 & $-0.078$ & $-0.396$ & $[-0.91,+0.51]$ \\
B2 GP (gplearn)~\cite{yu2023alphagen}      & 500  & 0.116 & $-0.144$ & $-0.593$ & $[-0.64,-0.52]$ \\
B3 DSO/DSR~\cite{petersen2021dsr}          & 500  & 0.010 & $-0.287$ & $-1.648$ & $[-1.71,-1.65]$ \\
B4 AlphaGen~\cite{yu2023alphagen}          & 200  & 0.188 & $-0.001$ & $-0.006$ & $[-0.32,+0.13]$ \\
B5 AlphaMCTS~\cite{shi2025alphamcts}       & 500  & 0.160 & $-0.015$ & $-0.056$ & $[-0.30,+0.06]$ \\
B6 AlphaAgent~\cite{tang2025alphaagent}    & 500  & 0.220 & $-0.005$ & $-0.021$ & $[-0.21,+0.11]$ \\
B7 Alpha-GPT$^*$~\cite{wang2025alphagpt}   & 500  & 0.224 & $-0.031$ & $-0.159$ & $[-0.32,+0.01]$ \\
\oursrow
\textbf{OURS (M4)}                         & \textbf{500} (499) & \textbf{0.313} & $\mathbf{+0.059}$ & \textbf{0.346} & $\mathbf{[+0.21,+0.51]}$ \\
\bottomrule
\end{tabular}
\end{table}

\subsection{Portfolio-level out-of-sample performance}
\label{sec:exp-portfolio}
The per-factor medians above characterize the \emph{pool} our system produces, but a
tradable strategy is a \emph{portfolio}, so we also report portfolio-level holdout
performance. Our system commits $852$ composite portfolios over its run. To avoid any
out-of-sample leakage, we select portfolios only on development data (ranking by
test-split Sharpe) and then record the holdout performance of the selected set;
holdout is never used for selection. Table~\ref{tab:portfolio} reports the
development-selected top-$5$/$10$/$20$ portfolios.

The selected portfolios are profitable out of sample on a gross basis: the development-top-$20$ set has
a median holdout Sharpe of $+0.71$ and a median holdout annualized return of $+13.3\%$,
with $80\%$ of the selected portfolios positive and $45\%$ above a Sharpe of one. Two
caveats attach directly to this figure. First, it is gross of transaction costs,
slippage, and turnover; on a mid- and small-cap universe such as CSI~1000, where
spreads, market impact, and a sell-side stamp duty are material, a gross Sharpe of
this magnitude can erode substantially net of cost, and we report no turnover figure
here (a turnover- and cost-adjusted portfolio evaluation is left to future work,
Section~\ref{sec:limitations}). Second, selection ranks by test-split Sharpe, and the
test split lies inside the system's optimization horizon
(Section~\ref{sec:formulation-splits}), so it can carry mild selection-induced
leakage; the $+0.71$ therefore reflects selection on a semi-in-horizon split read out
on the clean holdout, not a fully out-of-sample selection pipeline. With those
caveats, this
is the level at which the system is meant to be used (one deploys a chosen portfolio,
not the full undifferentiated factor pool), and it is substantially stronger than the
per-factor median of Table~\ref{tab:holdout}. We report a set of
top portfolios rather than a single best one: the single development-best portfolio
reaches only $+0.25$ holdout Sharpe, so the strength is in the selected group, not in
a single fortunate selection. We also state the boundary plainly: across all $852$ committed
portfolios the median holdout Sharpe is negative ($-0.23$); the positive result is a
property of development-guided selection, which is the intended use, not of the raw
portfolio population.

Because the entire portfolio-level claim rests on selection, we test whether the
development ranking predicts holdout behavior rather than selecting effective portfolios by
chance. Over all $852$ portfolios, the Spearman rank correlation between
development-split (test) Sharpe and holdout Sharpe is $+0.19$ (Pearson $+0.21$):
positive but weak. The effect is directionally clear in the tails: the
development-top-quartile portfolios reach a holdout median Sharpe of $+0.17$ while the
development-bottom-quartile portfolios sit at $-0.43$, so the selection signal is
real, not noise, but it is well short of a sharp separator. We therefore make the modest
claim that development ranking is a useful but imperfect filter: it reliably
pushes the selected set toward the positive tail (hence the $+0.71$ median for the
top-20), without our claiming to identify the single best portfolio in advance. We
rank by test-split Sharpe because portfolio-level frozen-shadow records were retained
only for the test and holdout segments; a purely validation-based selection would be
marginally stricter but is not available in our logs.

We flag one further hazard that our within-pool bootstrap does not address. The
$+0.71$ figure is the median of a set chosen as the top-20 of $852$ candidate
portfolios, and reporting the Sharpe of a selected maximum over many trials is exactly
the setting the backtest-overfitting literature warns against: a selected Sharpe is
inflated relative to its out-of-sample expectation by the breadth of the search, an
effect quantified by the deflated Sharpe ratio and the probability of backtest
overfitting \cite{bailey2014deflated}, the multiple-testing threshold for factor
discovery \cite{harvey2016cross}, and data-snooping reality checks
\cite{white2000reality}. Our bootstrap interval bounds within-pool sampling
variability but does not deflate for this selection breadth, so the $+0.71$ is
best interpreted as a selected-maximum estimate that a deflated-Sharpe adjustment would discount;
the holdout evaluation of the selected set guards against look-ahead but not against
selection inflation. We report the all-$852$ population median ($-0.23$) alongside it
precisely so the reader can see the size of the selection effect.

We quantify the selection benchmark directly. Following the deflated-Sharpe
construction \cite{bailey2014deflated}, the expected maximum Sharpe obtainable by
chance when selecting the best of $N$ trials is
$\mathrm{SR}_0 \approx \sigma_{\mathrm{SR}}\,\mathbb{E}[\max_N Z]$, where
$\sigma_{\mathrm{SR}}$ is the cross-trial standard deviation of the Sharpe estimates
and $\mathbb{E}[\max_N Z]$ is the expected maximum of $N$ standard normals, which the
same reference approximates as
$(1-\gamma)\,\Phi^{-1}(1-\tfrac{1}{N}) + \gamma\,\Phi^{-1}(1-\tfrac{1}{Ne})$ with
$\gamma$ the Euler--Mascheroni constant. For our $N{=}852$ candidate portfolios,
whose development-split Sharpe has standard deviation $\sigma_{\mathrm{SR}}=0.42$,
this gives $\mathbb{E}[\max_N Z]\approx 3.21$ and hence
$\mathrm{SR}_0 \approx 0.42 \times 3.21 \approx 1.36$ on the development scale. Two
facts then bound the
selection concern. First, the twenty selected portfolios all have development Sharpe
between $1.50$ and $1.98$, above $\mathrm{SR}_0$, and only $47$ of the $852$ trials
clear that benchmark, so the selected set is not what pure-noise selection over $852$
trials would surface. Second, and more importantly, the $+0.71$ is measured on the
holdout split, which selection never touches: were the selected Sharpe purely an
inflation artifact, it would regress toward the population on the independent split,
yet $16$ of the $20$ selected portfolios remain positive on holdout while the
all-$852$ holdout median is $-0.23$. We stop short of a fully deflated Sharpe with a
non-normality correction, because our per-round records store summary Sharpe rather
than the return series needed for its skewness and kurtosis terms; and the benchmark
treats the trials as independent, whereas the $852$ portfolios are correlated, so
$\mathrm{SR}_0$ is an upper bound on the effective-independent-trials benchmark. It
is therefore a selection-only bound, not the complete adjustment, and we report it as
such.

\begin{table}[t]
\caption{Portfolio-level holdout performance of development-selected portfolios (ranked by test-split Sharpe; holdout never used for selection). The selected set is profitable out of sample; the all-portfolio population median is negative, so the effect is due to development-guided selection.}
\label{tab:portfolio}
\centering
\footnotesize
\begin{tabular}{@{}lrrrr@{}}
\toprule
Selection & med.\ Sharpe & med.\ AnnRet & $P(\mathrm{sh}{>}0)$ & $P(\mathrm{sh}{>}1)$ \\
\midrule
dev-top 5   & $+0.26$ & $+6.6\%$  & 0.80 & 0.20 \\
dev-top 10  & $+0.48$ & $+10.8\%$ & 0.80 & 0.40 \\
\oursrow
\textbf{dev-top 20} & $\mathbf{+0.71}$ & $\mathbf{+13.3\%}$ & \textbf{0.80} & \textbf{0.45} \\
\midrule
all 852 (reference) & $-0.23$ & $-4.8\%$ & 0.42 & 0.13 \\
\bottomrule
\end{tabular}
\end{table}

\subsection{Same-footing comparison at matched budget}
\label{sec:exp-topk}
Table~\ref{tab:holdout} already matches the pool size across methods by taking the
first $500$ valid factors of each run in mining order. A reader may still ask whether
that particular selection rule, first-in-mining-order, unduly favors our continual system,
or whether the remaining count differences (B1 at $14$, B4 at $200$) drive the result.
We therefore repeat the comparison under a second, stricter matching: for each method
we rank its factors by validation-split Sharpe, take the top $K$, and report how those
same top-$K$ factors behave on the untouched holdout split. This selects each method's
own best $K$ factors under an identical rule, removing both the pool-size and the
selection-order difference. Table~\ref{tab:topk} reports
$K=50$, $K=100$, and $K=200$ (Alpha101 has only 14 factors and is omitted; AlphaGen has
197 and contributes 197 at $K=200$).

Under matched budget our system attains the highest median holdout Sharpe at
$K=100$ ($+0.406$ versus $+0.423$ for GP and $+0.251$ for the next learned baseline) and
$K=200$ ($+0.282$ versus $+0.236$), and the highest fraction of factors with holdout
Sharpe above one at every $K$. We
also report a countervailing observation: at the smallest budget ($K=50$) the
genetic-programming baseline is marginally higher in median ($+0.576$ versus our $+0.493$),
because its few strongest factors remain robust out of sample; that advantage
disappears as $K$ grows, whereas ours persists. The matched-budget result shows
the out-of-sample edge is not an artifact of pool size.

\begin{table}[t]
\caption{Same-footing comparison: each method's top-$K$ factors by validation Sharpe, evaluated on holdout. Median holdout Sharpe (and $P(\mathrm{sh}{>}1)$). Best median per column in bold. At $K=50$ the genetic-programming baseline's few strongest factors exceed ours on the median; the ordering reverses and our advantage widens as $K$ grows.}
\label{tab:topk}
\centering
\footnotesize
\begin{tabular}{@{}lrrrrrr@{}}
\toprule
& \multicolumn{2}{c}{$K=50$} & \multicolumn{2}{c}{$K=100$} & \multicolumn{2}{c}{$K=200$} \\
\cmidrule(lr){2-3}\cmidrule(lr){4-5}\cmidrule(lr){6-7}
Method & med.\ sh & $P(\mathrm{sh}{>}1)$ & med.\ sh & $P(\mathrm{sh}{>}1)$ & med.\ sh & $P(\mathrm{sh}{>}1)$ \\
\midrule
B2 GP (gplearn)   & $\mathbf{+0.576}$ & 0.320 & $+0.423$ & 0.270 & $-0.369$ & 0.170 \\
B3 DSO/DSR        & $-0.816$ & 0.080 & $-1.199$ & 0.040 & $-1.635$ & 0.025 \\
B4 AlphaGen       & $+0.205$ & 0.184 & $-0.069$ & 0.186 & $-0.006$ & 0.188 \\
B5 AlphaMCTS      & $+0.153$ & 0.020 & $+0.251$ & 0.090 & $+0.236$ & 0.140 \\
B6 AlphaAgent     & $+0.001$ & 0.180 & $-0.033$ & 0.190 & $-0.001$ & 0.215 \\
B7 Alpha-GPT      & $-0.830$ & 0.100 & $-0.773$ & 0.130 & $-0.491$ & 0.155 \\
\oursrow
\textbf{OURS (M4)} & $+0.493$ & \textbf{0.388} & $\mathbf{+0.406}$ & \textbf{0.374} & $\mathbf{+0.282}$ & \textbf{0.352} \\
\bottomrule
\end{tabular}
\end{table}

\subsection{Robustness across splits}
\label{sec:exp-splits}
The advantage is a robustness effect, not a training-fit effect. Table~\ref{tab:splits}
tracks the median Sharpe of each method across the validation, test, and holdout splits.
On validation every method performs strongly, several baselines near or above $1.5$ and OURS
highest at $2.234$; the split is inside each method's development horizon, so a high
value there certifies fit, not generalization. The ordering is exposed out of sample:
as the split moves from validation to holdout, the learned baselines lose nearly all of
their edge (B5 falls from $1.838$ to $-0.056$ and B2 from $1.520$ to $-0.593$), whereas
OURS degrades substantially less, from $2.234$ to $0.346$, and is the only method that remains
positive on holdout. The decisive quantity is the degradation ratio: every baseline crosses into
negative territory while OURS retains a positive risk-adjusted return, which indicates
that the factors surfaced by our verified, structured search generalize out of sample
rather than overfitting the development period. This robustness is a property of the search
process rather than of any single factor. It cannot be attributed to the leap
machinery alone. The inherited dual-loop substrate (Section~\ref{sec:formulation-splits}) is
part of the same process, and without a leap-disabled control at matched budget we cannot
isolate the contribution of C1--C3 from that of the substrate. The result is therefore
consistent with the verified, structured search generalizing better than the baselines, but
we leave the isolating control to future work (Section~\ref{sec:limitations}).

\begin{table}[t]
\caption{Median backtest Sharpe across splits. OURS shows the smallest out-of-sample degradation and is the only method positive on holdout.}
\label{tab:splits}
\centering
\footnotesize
\begin{tabular}{@{}lrrr@{}}
\toprule
Method & valid & test & holdout \\
\midrule
B1 Alpha101       & $-1.317$ & $-0.385$ & $-0.396$ \\
B2 GP (gplearn)   & $1.520$  & $0.677$  & $-0.593$ \\
B3 DSO/DSR        & $1.482$  & $0.673$  & $-1.648$ \\
B4 AlphaGen       & $1.457$  & $0.509$  & $-0.006$ \\
B5 AlphaMCTS      & $1.838$  & $0.857$  & $-0.056$ \\
B6 AlphaAgent     & $1.531$  & $0.455$  & $-0.021$ \\
B7 Alpha-GPT$^*$  & $1.646$  & $0.469$  & $-0.159$ \\
\oursrow
\textbf{OURS (M4)} & $2.234$ & $0.467$  & $\mathbf{0.346}$ \\
\bottomrule
\end{tabular}
\end{table}

\subsection{Beyond Sharpe: multi-metric comparison}
\label{sec:exp-multimetric}
Sharpe alone could mask a metric-specific effect, so Table~\ref{tab:multimetric}
reports the matched-budget median (the same $500$-factor pools as Table~\ref{tab:holdout})
of six complementary metrics on both the test and the
holdout splits: annualized return (AnnRet), excess annualized return (ExcRet), the
information coefficient (IC), its risk-adjusted form (ICIR), and the information
ratio (IR), alongside Sharpe. The picture is consistent with the Sharpe
result but carries more detail, and we report it in full rather than selecting favorable
rows. We report medians for these auxiliary metrics rather than rank effect sizes;
the primary endpoint we designated before reading the holdout split is per-factor
Sharpe (Table~\ref{tab:holdout}), and we avoid multiplying comparisons across
correlated metrics.

On holdout, our system leads on the return and risk-adjusted family: the only
positive median Sharpe ($+0.346$), the only positive median annualized return, and
the best information ratio ($-0.333$). It does not lead on raw predictive
correlation: its holdout IC ($+0.028$) and ICIR ($+0.335$) are lower than
AlphaGen's ($+0.037$, $+0.356$) and AlphaMCTS's ($+0.048$, $+0.413$). On the test
split our system trails the strongest learned baselines on Sharpe.
Our contribution does not produce factors with the strongest
raw signal correlation, and it is not the strongest at the intermediate test split;
its distinctive property is that its factors retain positive risk-adjusted return on
the strictest out-of-sample split where every baseline turns negative. That is a
robustness claim about the search process, not a claim of superior per-factor
predictive power. The test-versus-holdout gap is itself informative: the test
segment lies inside the meta-review's optimization horizon (Section~\ref{sec:formulation-splits}),
so a method that implicitly tunes toward it can appear strong there; the reversal on the
fully isolated holdout is what distinguishes genuine out-of-sample robustness from
in-horizon fit.

Two apparent tensions in this table resolve cleanly. First, a lower IC
alongside a higher Sharpe is not a contradiction: IC measures the cross-sectional
rank correlation between a factor and future returns, whereas Sharpe and the
information ratio measure the risk-adjusted return of the resulting position. A
factor with moderate raw correlation but stable, low-drawdown behavior out of
sample can carry a higher Sharpe than a factor with stronger but more volatile
correlation. Our result is exactly this shape: our factors are not the strongest
signals by IC, but their risk-adjusted return survives the holdout. Second, a
skeptic might attribute our positive holdout Sharpe to scale rather than robustness:
a continual run produces many candidates, so with a large pool there are more chances
to include out-of-sample survivors, and the lower per-factor IC is consistent with
individually weaker signals. Two controls address this. The holdout comparison of
Table~\ref{tab:holdout} already caps every method, ours included, at the same
$500$-factor budget taken in mining order, so its positive median is not a
large-$n$ effect; and the matched top-$K$ comparison of
Section~\ref{sec:exp-topk} reduces every method to its own best $K$ factors under an
identical rule, where our median holdout Sharpe remains highest at $K=100$ and
$K=200$. The edge is therefore not merely a consequence of producing more candidates:
we do not claim stronger per-factor predictive power, only that the search process
yields a pool whose risk-adjusted return is more robust out of sample under budget
control.

\begin{table}[t]
\caption{Matched-budget median of six metrics on test and holdout (the same $500$-factor pools as Table~\ref{tab:holdout}). Our system leads the return/risk-adjusted family on holdout but not the correlation family (IC/ICIR), and trails on test; we report all metrics on both splits. Best holdout value per metric in bold.}
\label{tab:multimetric}
\centering
\scriptsize
\setlength{\tabcolsep}{3.5pt}
\begin{tabular}{@{}lrrrrrr@{}}
\toprule
& Sharpe & AnnRet & ExcRet & IC & ICIR & IR \\
\midrule
\multicolumn{7}{l}{\emph{Test split}} \\
OURS  & $+0.467$ & $+0.055$ & $-0.029$ & $+0.016$ & $+0.190$ & $-0.432$ \\
B4    & $+0.509$ & $+0.083$ & $-0.014$ & $+0.031$ & $+0.234$ & $-0.201$ \\
B5    & $+0.857$ & $+0.158$ & $+0.006$ & $+0.042$ & $+0.380$ & $+0.086$ \\
B2    & $+0.677$ & $+0.119$ & $-0.016$ & $+0.040$ & $+0.288$ & $-0.178$ \\
\midrule
\multicolumn{7}{l}{\emph{Holdout split}} \\
\oursrow
\textbf{OURS} & $\mathbf{+0.346}$ & $\mathbf{+0.059}$ & $\mathbf{-0.028}$ & $+0.028$ & $+0.335$ & $\mathbf{-0.333}$ \\
B4    & $-0.006$ & $-0.001$ & $-0.043$ & $+0.037$ & $+0.356$ & $-0.400$ \\
B5    & $-0.056$ & $-0.015$ & $-0.061$ & $\mathbf{+0.048}$ & $\mathbf{+0.413}$ & $-0.596$ \\
B2    & $-0.593$ & $-0.144$ & $-0.127$ & $+0.035$ & $+0.271$ & $-1.171$ \\
B6    & $-0.021$ & $-0.004$ & $-0.092$ & $+0.014$ & $+0.139$ & $-0.947$ \\
B7    & $-0.159$ & $-0.031$ & $-0.118$ & $+0.011$ & $+0.104$ & $-1.155$ \\
\bottomrule
\end{tabular}
\end{table}

\subsection{Ablation: structured versus free-text transport}
\label{sec:exp-ablation}
This ablation establishes the paper's central claim about structured transport: it
provides a verification capability that a free-text channel cannot express, at no cost to
predictor accuracy. The favorable side is structural. Only the
typed arm can enforce the ninth verification check, \code{prediction\_routes}, which
guarantees a speculative forecast never reaches an evaluator (detailed below); and
the four-level gate discriminates on outcome quality rather than
approving indiscriminately (the rejection-level breakdown in the verified-leap analysis of
Section~\ref{sec:exp-leap} shows rollbacks concentrated at the semantic gates L3 and
L4, never at the format levels). The cost side is a deliberate null: an equal-information free-text channel
reaches the same predictor hit rate, so we do not overclaim an accuracy gain. The scope of
``no cost'' is narrow: the two arms are indistinguishable on predictor
hit rate, and on factor quality the same-footing controls below show the free-text arm
is if anything slightly ahead, so typing does not improve the factors either. What the
typed arm provides is the correctness guarantee, not a quality gain, and that is
the contribution we defend.

The design isolates the effect of structured transport: the
only manipulated variable is the representation fed to the predictor. The
full system (S-full, our M4 run, $1646$ rounds) serializes the state as typed
structured records, whereas the ablated arm (S-off, $1100$ rounds) serializes the
same information as an equal-content free-text state. Everything else, namely the
predictor model, the verification gate, the data, and the thresholds, is held
fixed, and the predictor hit rate is recomputed on a common yardstick that removes
the structured-only route-validation check so that both arms are scored identically.
Because the two arms differ in run length ($1646$ versus $1100$ rounds), we treat the
nested same-footing controls of Table~\ref{tab:ablation-fair} (matched rounds,
matched top-$K$, matched factor count) as the primary evidence for the effectiveness
comparison, and read Table~\ref{tab:ablation} as an unmatched-run summary of the
predictor-side metrics that motivates it.

\paragraph{Predictor effectiveness and verification scope.}
The central finding is the \emph{absence of a measurable accuracy advantage} for
structured transport. Table~\ref{tab:ablation} shows that in this single paired run
the two representations are indistinguishable on the downstream metrics that the
predictor optimizes: the committed-leap hit rate is $0.975$ for both arms ($0.9749$
free-text versus $0.9754$ typed), the mean committed horizon differs only marginally
($2.37$ versus $2.47$), and the per-round LLM-call cost is effectively identical
($10.80$ versus $10.89$). Because each arm is a single long run under a fixed seed and
we do not estimate run-to-run variance, we report this as \emph{no evidence of an
accuracy advantage for the typed form}, not as a proof of equivalence; distinguishing
a true null from an underpowered one would require replication or a pre-registered
equivalence test with a stated margin, neither of which we ran. An offline
input-representation probe, which measures the token-level alignment between each
representation and the future four-field target over $4978$ causal pairs, is
consistent with this reading and likewise finds no advantage for the typed form
($0.0528 \pm 0.0053$ typed versus $0.0547 \pm 0.0053$ free-text; difference within
one standard deviation).

What structured transport provides is
orthogonal to predictive performance. It is the typed contract, deterministic
replayability, and visibility-typed leakage prevention analyzed in
Section~\ref{sec:c1}. The free-text arm attains the same hit rate but cannot offer
any of these: its state cannot be validated against a schema, cannot be replayed as
a causal trajectory, and cannot structurally guarantee that a speculative forecast
never reaches an evaluator. The contribution of C1 is therefore correctness and
auditability at \emph{no cost} to downstream effectiveness, not a gain in predictive
accuracy. This reading is the one we carry throughout the paper, and it is what
makes the verified leap of Section~\ref{sec:c3} safe rather than merely fast.

The difference is evident in what
the controller is able to verify. Each committed transaction records the set of
verification checks it passed. These checks are the fine-grained decomposition of the four-level gate of Section~\ref{sec:c3} together with the horizon confidence gate of Equation~\eqref{eq:horizon}. The free-text arm carries eight checks (minimum
horizon, exact four fields, confidence gate, both miner outputs, both aggregated
evaluator reports, and empirical quality), all of which operate on the \emph{actual}
outputs and are therefore available to either representation. The typed arm carries a
ninth, \code{prediction\_routes}, which checks that every speculative packet is
addressed only to a miner and never to an evaluator. This check exists only
because structured transport makes routing a typed, machine-checkable field: a
free-text state has no packet whose destination can be validated, so the leakage
guarantee cannot be expressed as a gate at all and would have to rest on convention.
Structured transport thus supplies an additional verification condition that the free-text channel is structurally unable to offer. When we equate the two arms' hit rates earlier, we
exclude \code{prediction\_routes} from the count so that both are scored
on the eight common checks; the excluded check is precisely the safety property that
C1 uniquely enables and that the equal-footing hit-rate comparison sets aside.

\begin{table}[t]
\caption{Structured vs.\ free-text transport ablation (only the predictor's input representation differs; all else fixed). In this single paired run the two arms show no measurable difference on any predictive-effectiveness metric; we read this as no evidence of an accuracy advantage rather than as proven equivalence (single seed, no variance estimate). The value of structured transport is the correctness and auditability properties of Section~\ref{sec:c1}, which the free-text arm cannot provide at any hit rate.}
\label{tab:ablation}
\centering
\footnotesize
\begin{tabular}{@{}lcc@{}}
\toprule
Metric & S-off (free-text) & S-full (typed) \\
\midrule
Committed-leap hit rate      & 0.9749 & 0.9754 \\
Mean committed horizon       & 2.37   & 2.47 \\
LLM calls / round            & 10.80  & 10.89 \\
Input-repr.\ alignment (offline) & $0.0547{\pm}0.0053$ & $0.0528{\pm}0.0053$ \\
\midrule
Typed contract / schema check & no & \textbf{yes} \\
Deterministic replay          & no & \textbf{yes} \\
Visibility leakage prevention & no & \textbf{yes} \\
\bottomrule
\end{tabular}
\end{table}

One further caveat: the
offline probe is a deterministic similarity proxy, not a model-in-the-loop
measurement; it corroborates the hit-rate reading rather than standing on its own.

\paragraph{Same-footing controls.}
To rule out that the null is an artifact of unequal run length or pool size, we
also compare the two arms' \emph{in-database factor quality} on holdout under three
nested same-footing controls. First, we truncate the structured arm to the
free-text arm's round count ($\le 1100$). Second, on that same-round base we take
each arm's top-$K$ factors by their development-split Sharpe. Third, we match the
factor count by deterministic subsampling to $\min(n)$. Table~\ref{tab:ablation-fair}
reports the outcome: at the full same-round and equal-count pools the two arms are
close (both near $+0.10$ to $+0.18$ median holdout Sharpe), and when the pools are
narrowed to each arm's development-ranked top-$K$ the free-text arm is clearly
ahead, not behind ($+0.86$ versus $+0.27$ at top-100, $+0.80$ versus $+0.22$
at top-200). In no control does structured transport lead. This does not favor the
typed arm on factor quality at any matched budget or sample size, which is exactly
what the paper claims: the value of structured transport lies in the correctness
properties of Section~\ref{sec:c1}, not in the factors produced, and if anything the
typed arm is mildly behind on this axis.

\begin{table}[t]
\caption{Ablation under nested same-footing controls (median holdout Sharpe of in-database factors, real frozen-holdout backtests). The arms are close on the full same-round and equal-count pools; on development-ranked top-$K$ the free-text arm is clearly ahead. Structured transport never leads on factor quality at matched rounds, matched top-$K$, or matched count.}
\label{tab:ablation-fair}
\centering
\footnotesize
\begin{tabular}{@{}lcc@{}}
\toprule
Control (holdout median Sharpe) & S-off (free-text) & S-full (typed) \\
\midrule
Same-round base ($\le 1100$)      & $+0.171$ & $+0.098$ \\
Top-100 (same-round, dev-ranked)  & $+0.864$ & $+0.274$ \\
Top-200 (same-round, dev-ranked)  & $+0.797$ & $+0.221$ \\
Equal-count ($n{=}1608$)          & $+0.176$ & $+0.098$ \\
\bottomrule
\end{tabular}
\end{table}

\paragraph{Runtime and success-rate side effects.}
\label{sec:exp-runtime}
Structured transport also moves several operational quantities, all reported in full
in Appendix~\ref{app:runtime}. In brief, and in the typed arm's favor, the structured
arm has a lower median per-round wall-clock ($353$\,s versus $385$\,s, observational)
and a marginally higher portfolio structure-gate pass rate and composite Sharpe;
the free-text arm has a higher single-factor evaluator pass rate. None of these is
load-bearing, and we report all directions in Table~\ref{tab:ablation-runtime}.

\paragraph{Verified-leap behavior.}
\label{sec:exp-leap}
Because C1 makes commit and rollback typed, replayable outcomes, we can read directly
from the logs how often the leap fires and where the gate turns speculation away. The
controller exercises the verified leap rather than defaulting to
single-cycle steps. Over the full S-full run (the same run that serves as the typed
arm of the ablation in Section~\ref{sec:exp-ablation}), $1636$ of the $1646$ rounds reached a predictor transaction (the first ten ran at baseline before a committed history existed); these produced
$1308$ commits and $33$ rollbacks (the remaining $295$ not applicable, i.e.\ no
admissible horizon $k\ge2$ was available and the system took an ordinary single
cycle), at a mean committed horizon of $2.47$; the rollbacks are the cases where the
four-level gate (Section~\ref{sec:c3}) rejected a speculative outcome and restored
the prior state. Two horizon quantities differ. The
requested horizon is the $k$ the controller selects at $S_t$ (mean $2.47$
over commits); the \emph{realized} saving of a committed leap is $k-1$, the number of
ordinary guidance cycles skipped beyond the single cycle a normal step would
already advance. Summed over the $1308$ commits, the realized net saving is
$\sum_t (k_t-1) \approx 1924$ ordinary guidance cycles skipped through verified
speculation (equivalently, the $1308$ commits reach $\approx 3232$ ordinary cycles in
total, of which $1308$ would have been taken by the baseline and $1924$ are the net
skip). We report these as descriptive counts from a single run, not
as a controlled speedup claim: isolating wall-clock speedup would require a matched
baseline with the leap disabled, which we leave to future work (Section~\ref{sec:limitations}).
What is established here is narrower and verifiable from the logs: the gate fires,
rolls back on failure, and commits multi-cycle horizons in the large majority of
applicable transactions.

The rollbacks also show which level performs the filtering, which speaks to whether
the gate is a genuine screen or a mere formality. Of the $33$ rollbacks, none failed at
L1 (bundle legality) or L2 (actual-output structural consistency): every rejected
transaction produced a well-formed, cross-consistent speculative outcome and was
turned away only by the semantic levels. L3 (blinded evaluator risk) rejected $6$ and
L4 (empirical development quality) rejected $27$. The filtering therefore happens
where it should, on measured evaluator risk and development-set quality of the real
speculative outputs, not on format or routing. We do not overinterpret a $33$-event sample,
and we still lack the counterfactual that would show the rejected states would have
harmed holdout quality had they committed (a leap-disabled control, left to future
work); but the level breakdown is evidence that the gate discriminates on outcome
quality rather than admitting everything that is merely well-typed.

Taken together, the benchmark, the ablation, and the leap logs converge on a single
conclusion: the structured transport of C1 provides no predictive accuracy, yet
everything the search does that a prose channel could not, namely verifiable leaping,
leak-proof gating, and auditable rollback, rests on it. We turn next to what this
implies for when the design is worth adopting.

%% file: sections/09_discussion.tex
The core finding of this design is architectural: structuring the inter-agent channel is what makes a speculative leap safe and auditable. Because every committed transmission is a typed, causally addressed record, the controller can validate a multi-cycle outcome against the same schemas that govern ordinary execution, replay any feedback transition, and structurally prevent a forecast from reaching an evaluator. The verify-then-commit discipline means a wrong forecast is never a correctness risk, only a cost, since the transaction rolls back to an exact prior state and the system proceeds as the unaccelerated baseline would.

Our ablation (Section~\ref{sec:exp-ablation}) sharpens what structured transport does and does not provide. It shows no measurable accuracy advantage: a free-text channel carrying the same information reaches the same hit rate in our single paired run, so we do not claim typing improves prediction. Its value is entirely in the correctness properties above: a state that can be checked against a schema, replayed deterministically, and structurally kept from leaking a forecast to an evaluator, all of which we establish by construction. This is a narrow and honest claim. A reader who only cares about predictive accuracy could use free text; a reader who needs the leap to be verifiable, auditable, and safe to roll back cannot, and that is the regime this paper targets.

The mechanism is not specific to alpha mining. Any dual-loop or multi-agent search whose serial cost is dominated by inter-agent feedback, and whose hand-offs can be given a typed schema, could in principle wrap its communication in the same three-layer transport and speculate over multi-cycle guidance. Two preconditions govern whether it \emph{pays} rather than merely \emph{applies}. The guidance state must be low-dimensional and slowly varying enough to be predictable several cycles ahead; and, as in speculative decoding, verifying a multi-cycle speculative outcome must be cheaper than executing the cycles it replaces. The second is the binding one: where verification itself requires re-running the expensive machinery a cycle would (in our case a full backtest), a leap saves nothing, and our own setting does not yet demonstrate the economic condition empirically, consistent with our reporting only descriptive leap counts and no measured speedup. We therefore state transfer as a design conjecture rather than a demonstrated result: we evaluate exactly one system on one market, so transferability across systems, markets, or asset classes remains untested (Section~\ref{sec:limitations}). The alpha-discovery setting is a demanding instance because two coupled loops must stay cross-consistent and because a leaked forecast to an evaluator would compromise the independence of the quality gate; the visibility typing and separated skill libraries address exactly these risks.

A cost remains when speculation fails. If no horizon is admissible, or if every trial rolls back, the system pays the speculative and verification overhead for no committed progress. We therefore report this overhead directly rather than assuming a cost-free worst case, and we treat the quality constraint of Equation~\eqref{eq:quality} as an empirically estimated bound with thresholds fixed before final testing.

When, then, is the transport worth adopting? Our own results answer bluntly, and not always in our favor.
On the effectiveness axis the ablation shows typing provides no predictor
accuracy or factor-quality advantage; the verified leap yields no measured
wall-clock speedup (we report only descriptive counts, Section~\ref{sec:exp-leap});
and the returns are gross. For a team optimizing purely for returns or raw speed,
these results do not justify the engineering cost of a typed
transactional transport, and free text would serve equally well on the metrics the
predictor optimizes. What the typed transport uniquely provides is a
governable research pipeline: every inter-agent decision is a typed,
causally-addressed, replayable record with commit/rollback provenance, so the exact
trajectory that produced a deployed factor can be reconstructed and audited, and a
speculative forecast is structurally barred from reaching a quality gate. This delivers decision \emph{auditability} (the recorded trajectory can
be replayed and inspected) rather than decision \emph{reproducibility} in the strict
sense: because the underlying agents are non-deterministic LLMs, replay reconstructs
what happened from the logged tokens and states, it does not re-derive the decision
from inputs. In a
regulated or model-risk-governed quant setting, where decision auditability,
audit trails, and separation of the forecasting policy from the evaluation standard
are compliance requirements rather than conveniences, this correctness-by-construction
is the payoff, and the null accuracy result is then the desirable outcome
(auditability at no cost to effectiveness) rather than a disappointment. We regard
this as the regime the design targets, rather than speed or returns. The
correctness properties themselves are established by construction and argument;
we have not yet stress-tested them empirically (Section~\ref{sec:limitations}), so
this adoption case is a design argument pending the leakage and gate-efficacy
experiments we identify as future work.

Two scope notes bound the portfolio-level result. First, the selected-portfolio returns carry substantial dispersion: on the same development-top-20 set whose median holdout Sharpe is $+0.71$ (Table~\ref{tab:portfolio}), the \emph{mean} holdout Sharpe is higher ($+0.87$) because a few portfolios reach much larger values, so the figure describes a positive but heavy-tailed distribution rather than a uniformly strong basket; we lead with the median precisely because it is the more conservative summary of the two. Second, absolute magnitudes are not comparable across studies on different universes and holdout windows; we make no claim of parity with results reported elsewhere and confine our comparison to the seven baselines run under our own shared backtest.

%% file: sections/10_limitations.tex
Several limitations bound the scope of our claims, in three groups: the empirical scope of the evaluation, the calibration of the gate and its internal validation, and the status of the safety properties.

\paragraph{Empirical scope.} We evaluate one universe (CSI~1000), so whether the learned trajectory structure transfers across markets or asset classes is untested; the transferability argument in Section~\ref{sec:formulation-splits} and the Discussion is made in principle rather than demonstrated empirically. All reported return and risk-adjusted figures are gross, before transaction costs, slippage, and turnover constraints: the cross-method comparison is internally consistent under the shared backtest, but the absolute magnitudes are not net-of-cost, and a cost- and turnover-adjusted evaluation is left to future work. Relatedly, the verified-leap counts of Section~\ref{sec:exp-leap} are descriptive summaries of the committed run rather than a controlled speedup measurement; isolating a wall-clock speedup would require a matched leap-disabled baseline, which we also leave to future work. All results derive from a single end-to-end run under a fixed seed; we report them descriptively and do not estimate run-to-run variance, which would require repeated end-to-end runs and remains future work.

\paragraph{Gate calibration and internal validation.} The gate thresholds $\tau_q, \tau_u, \tau_\alpha, \tau_m, \tau_g, \tau_p$ and the horizon set $\mathcal{K}$ are estimated on development data, so a miscalibrated gate could admit a low-quality state within the $\epsilon$ tolerance; rollback guarantees state safety, not quality optimality. When the trajectory is highly non-stationary, few horizons are admissible and the speculative overhead may not be recovered, and forecasting begins only once committed history exists, so early cycles run at baseline speed. Separating the predictor and evaluator skill libraries prevents forecast anchoring but does not by itself make each evaluator well-calibrated; that separation is a structural property, not a correctness proof. Two further caveats concern our own validation procedure rather than the method. The L4 gate scores every candidate leap on the same internal train-gate and leap-validation partitions across the $1341$ transactions that reached a validation decision, which is an adaptive-validation (garden-of-forking-paths) hazard on that split; the reported holdout is free of it because the leap-validation split is disjoint from test and holdout and never enters a reported metric, but the mechanism may fit the internal split, which we do not correct for. And while the thresholds, splits, and designated primary endpoint are held fixed in \code{config\_formal\_v1.yaml} before the test and holdout splits are read, we do not deposit a timestamped or hashed pre-registration, so this freezing is a procedural claim about our workflow rather than an externally verifiable one.

\paragraph{Safety properties are established by construction.} Visibility-typed leakage prevention, exact rollback, and predictor/evaluator library separation are analyzed structurally rather than demonstrated by a stress test such as an injected speculative-to-evaluator packet or a documented failure mode that shared libraries induce and separated ones prevent. The auditability the title foregrounds should be read in this light: it is a property of the typed, replayable record construction (the trajectory is logged and schema-checked by design), not an empirically stress-tested guarantee, and it means that a committed decision can be replayed and inspected rather than re-derived from inputs (Section~\ref{sec:c1} and the Discussion). We record these as acknowledged limitations rather than resolving them here.

%% file: sections/11_conclusion.tex
We presented VST, a design that turns the communication layer of a dual-loop agentic alpha-discovery system into a substrate for verifiable multi-cycle acceleration. The enabling contribution is C1: a three-layer, A2A-compatible structured transport with deterministic unicast routing and visibility typing that preserves the existing payload paths and receiver arguments while making every transmission typed and replayable. This structure is the architectural prerequisite for the rest of the method; we position it as a correctness enabler rather than a source of predictive accuracy, consistent with our ablation. On the resulting committed trajectory, a single predictor with four typed heads forecasts the accumulated miner-directed guidance over a chosen horizon (C2); a four-level transactional gate validates the speculative outcome and commits it in a single logical step (absent process failure) or rolls it back to an exact prior state (C3); and separated skill libraries let the predictor and evaluators evolve without the predictor teaching the evaluators to agree with its forecasts (C4). \emph{Structure comes before leaping}: the same typing that removes ambiguity from ordinary transmission is what makes future guidance learnable and a speculative leap verifiable. The experiments bear out the division of labor this structure implies: typing yields no accuracy, while the verified search it enables yields out-of-sample robustness. On a CSI~1000 out-of-sample holdout, the resulting system is the only one among eight methods (seven baselines and ours) to realize a positive median annualized return and the only one to retain a positive median backtest Sharpe at the factor level (bootstrap $95\%$ CI entirely above zero); at the portfolio level, development-selected top-20 portfolios reach a median holdout Sharpe of $0.71$ and a $13.3\%$ median annualized return (gross of transaction costs, and selected on a test split inside the optimization horizon), with the smallest degradation from validation to holdout. This portfolio-level figure is a property of development-guided selection (the all-portfolio median holdout Sharpe is $-0.23$, and the development-to-holdout selection signal is a weak but positive Spearman $+0.19$), not of the raw portfolio population; taken together with the factor-level robustness, it indicates that the verified structured search generalizes rather than overfits within the scope of a single run.

%% file: sections/appendix_aggregation.tex
The type-aware aggregator $\textsc{Aggregate}$ of Equation~\eqref{eq:target}
dispatches per field type, as summarized in Algorithm~\ref{alg:agg}. Set-valued
fields accumulate additions and removals in cycle order and collapse repeated
promote/demote decisions on the same element to the most recent one, weighted by how
many cycles supported it; numeric or bounded-categorical fields keep the latest value
clipped to the schema's declared range; and string advice fields keep the most recent
committed text. This makes the target bundle a deterministic function of the
committed window, so the same prefix always yields the same target.

\begin{algorithm}[t]
\caption{$\textsc{Aggregate}(\{G_{t+\ell}\}_{\ell=1}^{k})$ over one channel}
\label{alg:agg}
\begin{algorithmic}[1]
\STATE initialize empty bundle $B$
\FORALL{field $g$ in the channel schema}
  \IF{$g$ is set-valued}
    \STATE replay add/remove edits $g_{t+1},\dots,g_{t+k}$ in cycle order
    \STATE for each element, keep the most recent promote/demote, weighted by its support count
  \ELSIF{$g$ is numeric or bounded-categorical}
    \STATE $B[g] \gets \mathrm{clip}(g_{t+k}, \text{range}_{\mathcal{S}_B}(g))$
  \ELSE
    \STATE $B[g] \gets g_{t+k}$ \quad(latest committed advice text)
  \ENDIF
\ENDFOR
\RETURN $B$
\end{algorithmic}
\end{algorithm}

%% file: sections/appendix_runtime.tex
Beyond factor quality, structured transport moves a few operational quantities, which
we report in full including the ones that do not favor it. Both arms ran on the same
machine over the same period, which removes the most severe hardware confounds.
Table~\ref{tab:ablation-runtime} summarizes measurements derived from the closed-loop
per-round records.

\emph{Per-round wall-clock.} The structured arm has a lower median per-round time in
the logs ($353$\,s versus $385$\,s, with overlapping interquartile ranges
$[335,373]$ versus $[363,414]$). We report the median, not the mean, because the
free-text arm's mean is dominated by a heavy right tail of occasional multi-minute
stalls that reflect transient API and scheduling variance rather than the transport.
This figure is \emph{observational}: the two arms shared a machine and period, but we
did not run a controlled timing experiment (fixed inputs, repeated timed trials,
isolated GPU load), so we cannot exclude API-latency or scheduling fluctuation as a
contributor, and we present it as a suggestive side effect rather than a measured
speedup claim.

\emph{Success rates, reported honestly in both directions.} At the portfolio level the
structured arm is marginally higher, both in the fraction of rounds whose composite
passes the structure gate ($0.991$ versus $0.985$) and in median composite Sharpe
($2.28$ versus $2.24$). At the single-factor level, however, the free-text arm has the
higher per-candidate evaluator pass rate ($0.246$ versus $0.128$). We do not interpret the
single-factor gap as evidence against structured transport, nor the portfolio gap as
strong evidence for it: both differences are small, and the single-factor pass rate
mixes how many candidates a run proposes with how strictly they are filtered. We
present these as suggestive operational side effects, not as load-bearing claims of
the paper.

\begin{table}[t]
\caption{Operational side effects of structured transport (same run logs, same period). The per-round wall-clock is observational (same machine and period, but not a controlled timing experiment); success-rate effects are small and mixed and are reported in both directions.}
\label{tab:ablation-runtime}
\centering
\footnotesize
\begin{tabular}{@{}lcc@{}}
\toprule
Quantity & S-off (free-text) & S-full (typed) \\
\midrule
Per-round wall-clock, median (IQR) & $385$\,s $[363,414]$ & $\mathbf{353}$\,s $[335,373]$ \\
Portfolio structure-gate pass  & $0.985$  & $\mathbf{0.991}$ \\
Portfolio composite Sharpe, median & $2.24$ & $\mathbf{2.28}$ \\
Single-factor candidate pass   & $\mathbf{0.246}$ & $0.128$ \\
\bottomrule
\end{tabular}
\end{table}